\documentclass[10pt,twocolumn,letterpaper]{article}

\usepackage[pagenumbers]{cvpr} 

\definecolor{cvprblue}{rgb}{0.21,0.49,0.74}
\usepackage[pagebackref,breaklinks,colorlinks,allcolors=cvprblue]{hyperref}

\usepackage{newfloat}
\usepackage{listings} 
\usepackage{adjustbox}
\usepackage{multirow}
\usepackage{booktabs}
\usepackage{pifont}
\usepackage{amsmath}
\usepackage{makecell}
\usepackage{color}
\usepackage[table,xcdraw]{xcolor}
\usepackage{enumitem}
\usepackage{algorithm}
\usepackage{algpseudocode}
\usepackage{longtable}

\def\paperID{} 
\def\confName{CVPR}
\def\confYear{2026}

\title{Evaluating Dataset Watermarking for Fine-tuning Traceability of Customized Diffusion Models: A Comprehensive Benchmark and Removal Approach}

\author{Xincheng Wang\\
Donghua University\\
Shanghai, China\\
\and
Hanchi Sun\\
Shanghai Jiao Tong University\\
Shanghai, China\\
}
\author{
Xincheng Wang\textsuperscript{1},
Hanchi Sun\textsuperscript{2},
Wenjun Sun\textsuperscript{3},
Kejun Xue\textsuperscript{1},
Wangqiu Zhou\textsuperscript{4},
Jianbo Zhang\textsuperscript{2},\\
Wei Sun\textsuperscript{5},
Dandan Zhu\textsuperscript{5},
Xiongkuo Min\textsuperscript{2},
Jun Jia\textsuperscript{2,*},
Zhijun Fang\textsuperscript{1,*},\\
\textsuperscript{1}Donghua University, \textsuperscript{2}Shanghai JiaoTong University,
\textsuperscript{3}Xidian University,\\
\textsuperscript{4}Hefei University of Technology,
\textsuperscript{5}East China Normal University\\
*Corresponding author(s)}
\begin{document}
\maketitle
\begin{abstract}
Recently, numerous fine-tuning techniques for diffusion models have been developed, enabling diffusion models to generate content that closely resembles a specific image set, such as specific facial identities and artistic styles. However, this advancement also poses potential security risks. The primary risk comes from copyright violations due to using public domain images without authorization to fine-tune diffusion models. Furthermore, if such models generate harmful content linked to the source images, tracing the origin of the fine-tuning data is crucial to clarify responsibility. To achieve fine-tuning traceability of customized diffusion models, dataset watermarking for diffusion model has been proposed, involving embedding imperceptible watermarks into images that require traceability. Notably, even after using the watermarked images to fine-tune diffusion models, the watermarks remain detectable in the generated outputs. However, existing dataset watermarking approaches lack a unified framework for performance evaluation, thereby limiting their effectiveness in practical scenarios. To address this gap, this paper first establishes a generalized threat model and subsequently introduces a comprehensive framework for evaluating dataset watermarking methods, comprising three dimensions: \textbf{Universality}, \textbf{Transmissibility}, and \textbf{Robustness}. Our evaluation results demonstrate that existing methods exhibit universality across diverse fine-tuning approaches and tasks, as well as transmissibility even when only a small proportion of watermarked images is used. In terms of robustness, existing methods show good performance against common image proces sing operations, but this does not match real-world threat scenarios. To address this issue, this paper proposes a practical watermark removal method that can completely remove dataset watermarks without affecting fine-tuning, revealing their vulnerabilities and pointing to a new challenge for future research.
\end{abstract}    
\section{Introduction}
\label{sec:intro}

\begin{figure*}[!t]
    \centering
    \includegraphics[width=.9\textwidth]{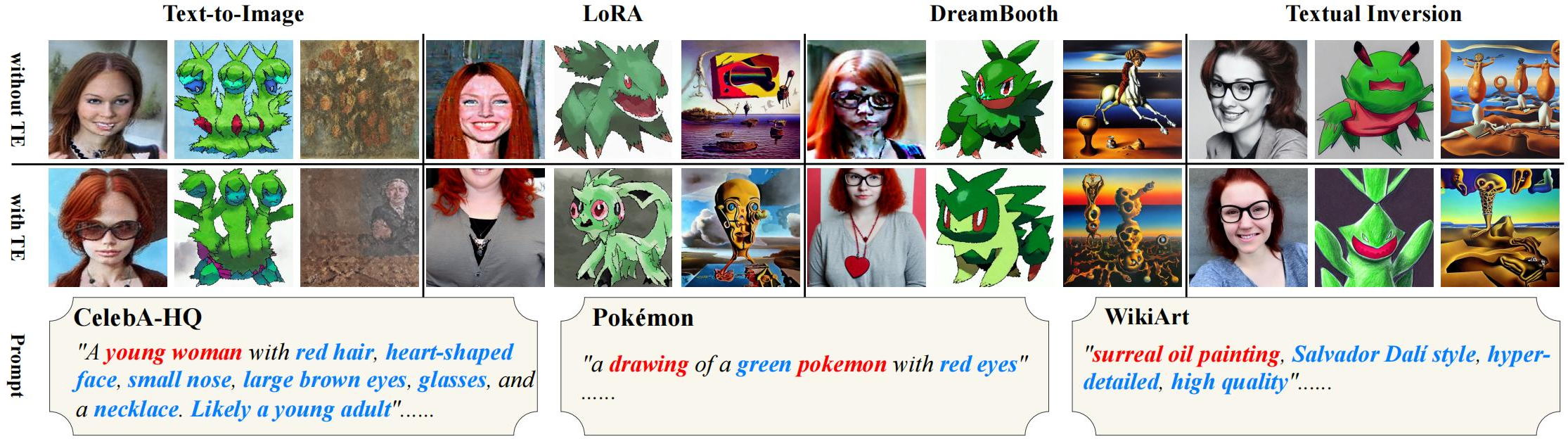}
    \caption{Visualization results of the four fine-tuning methods on three datasets. The first column shows the result without training the text encoder, the second column shows the result of training the text encoder, and the third column generates the corresponding prompt for each sample.}
    \label{fig:generated}
\end{figure*}

Currently, diffusion models have been widely adopted for customized content generation~\cite{sohl2015deep,song2020improved}  via fine-tuning techniques. These methods enable models to adapt to specific datasets, such as personal identities and image styles, and generate new images in different styles based on prompts, as shown in Figure~\ref{fig:generated}. While these fine-tuning methods significantly enhances the utility of diffusion models, it also raises serious security and ethical concerns. A primary issue is the risk of copyright infringement~\cite{duan2023diffusion,dubinski2025cdi}, particularly when proprietary images are utilized for fine-tuning without authorization. Furthermore, the generation of inappropriate or harmful content by fine-tuned models requires the implementation of source traceability mechanisms to establish clear accountability. To address this challenge, dataset watermarking technology~\cite{ren2024entruth,ding2024freecustom,wei2024powerful,ma2023generative} has been proposed to trace the outputs of diffusion models that have been fine-tuned using watermarked datasets. This technique involves embedding imperceptible watermarks into the images within a dataset intended for fine-tuning diffusion models. Notably, such watermarks are capable of persisting in the model's outputs following fine-tuning, thus facilitating post-hoc attribution. However, existing approaches~\cite{fernandez2023stable,wen2023tree,zhu2024watermark} exhibit varying definitions of threat models for dataset watermarking, which complicates the uniform evaluation of their performance in practical applications. Therefore, it is imperative to establish a unified evaluation framework to assess the performance of existing dataset watermarking techniques, thereby enabling the identification of currently optimal approaches and fostering the development of more practical watermarking solutions.

To address the limited understanding of adversarial behaviors in dataset watermarking for diffusion models, we introduce a universal threat model that formalizes realistic attack scenarios and serves as the foundation for systematic evaluation. Building upon this, we develop a unified evaluation framework encompassing three dimensions: \textbf{Universality}, measuring the applicability of dataset watermarking across diverse generation tasks and fine-tuning paradigms; \textbf{Transmissibility}, assessing the ability of watermarked subsets to propagate watermark signals through fine-tuning; and \textbf{Robustness}, quantifying resistance to post-processing and adversarial removal. We further construct a comprehensive benchmark to evaluate existing approaches under these criteria. Extensive experiments show that current methods perform well in universality and transmissibility but remain fragile when subjected to advanced watermark removal. To reveal this vulnerability, we propose a practical watermark removal attack that effectively erases dataset watermarks while maintaining model performance. Our findings highlight a critical gap in current designs and underscore the urgent need for more robust, adversary-aware dataset watermarking strategies for diffusion models. Our contributions can be summarized as follows:
\begin{itemize}
    \item This paper introduces a principled foundation for evaluating dataset watermarking in diffusion models by formalizing a universal threat model and proposing a unified framework that jointly measures universality, transmissibility, and robustness.
    \item This paper establishes a comprehensive benchmark for existing dataset watermarking methods based on the evaluation framework. Experiments across diverse generation tasks and various fine-tuning methods reveal the universality and transmissibility of existing methods.
    \item This paper proposes a practical watermark removal method to evaluate the robustness of existing dataset watermarking techniques. Experimental results show that current dataset watermarking methods are resilient to common image processing operations but vulnerable to targeted removal attacks.
\end{itemize}
\section{Related Works}
\label{sec:related}
\begin{figure*}[!t]
    \centering
    \includegraphics[width=.95\textwidth]{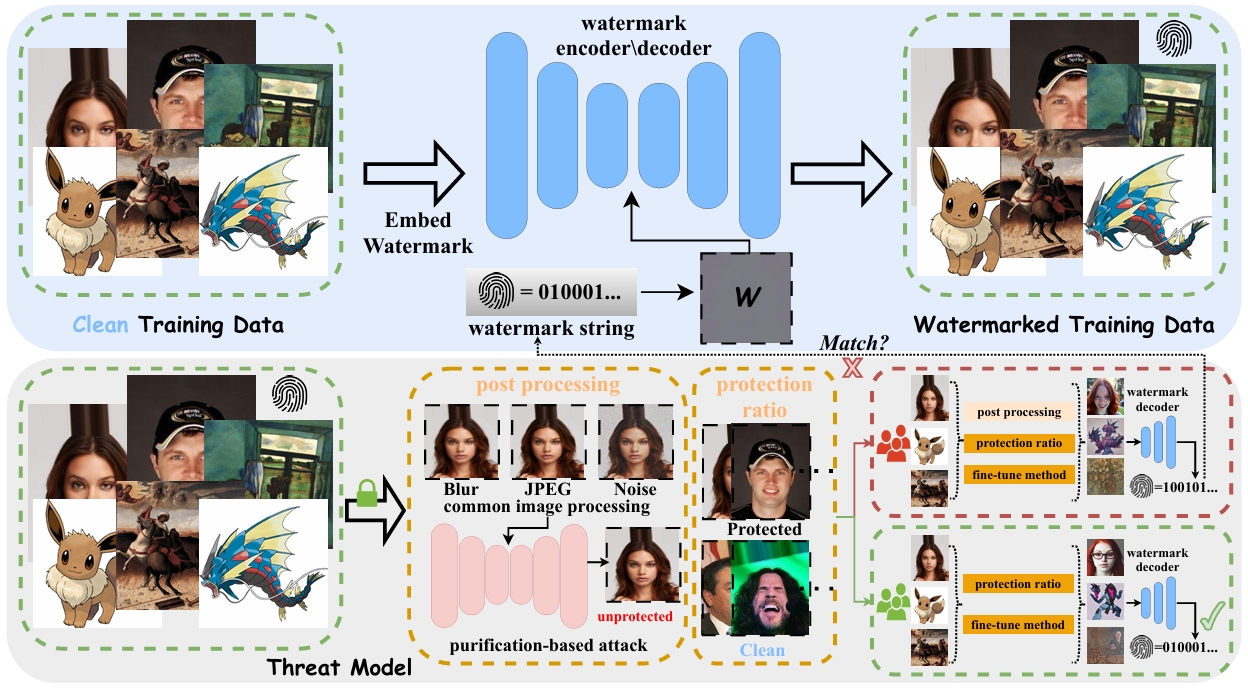}
    \caption{Overview of the threat model. \textbf{\textit{Image Owners}} embed binary watermarks into datasets to establish ownership and ensure traceability. Upon acquiring the data, \textbf{\textit{Image Users}} may generate customized images using various fine-tuning or model adaptation techniques. If the original watermark is successfully detected in the generated images, the protection mechanism is deemed effective; otherwise, it is considered to have failed.}
    \label{fig:threatmodel}
\end{figure*}
\subsection{Fine-tuning Stable Diffusion}
Due to the high computational cost of training stable diffusion models from scratch, recent research has focused on fine-tuning pre-trained models to add specific concepts. This approach leverages existing generative capabilities while greatly reducing training costs. Several fine-tuning methods have been proposed, such as Textual Inversion \cite{gal2022image}, DreamBooth \cite{ruiz2023dreambooth}, Custom Diffusion \cite{kumari2023multi}, Low-Rank Adaptation (LoRA) \cite{hu2022lora}, and Singular Value Diffusion (Svdiff) \cite{han2023svdiff}. These methods adapt pre-trained models in different ways to effectively introduce new concepts or styles. For instance, Textual Inversion only changes text embeddings, DreamBooth modifies the UNet architecture, Custom Diffusion targets cross-attention mechanisms, LoRA uses a low-rank matrix for parameter updates, and Svdiff adjusts singular values to create a compact parameter space.
\subsection{Image Watermarking}
Image watermarking refers to the process of embedding imperceptible information into carrier images, primarily for the purpose of asserting and verifying copyright ownership. Traditional watermarking techniques are typically classified into spatial domain and frequency domain methods \cite{cox2002digital,navas2008dwt,shih2003combinational}, where watermark data is embedded by modifying pixel intensities \cite{cox2002digital}, frequency coefficients \cite{navas2008dwt}, or a combination of both \cite{shih2003combinational,kumar2019review}. In recent years, an increasing number of digital watermarking approaches based on Deep Neural Networks (DNNs) \cite{zhu2018hidden,zhang2019invisible,weng2019high,tancik2020stegastamp} have been proposed, providing improved robustness and adaptability. Concurrently, several models have been developed to protect data copyrights from potential infringement by Generative Diffusion Models (GDMs). These techniques \cite{wang2024diagnosis,cui2025diffusionshield,zhao2023recipe,zhu2024watermark,li2025towards,yu2021artificial,cui2025ft} enable traceability of unauthorized data usage through the embedding of authorized encoding information or the application of image transformation strategies.

\section{Evaluation Framework}
\label{sec:eval_framework}
\subsection{Threat Model}
This section defines a universal threat model addressing  copyright protection and traceability of generated images in the context of fine-tuning diffusion models, as shown in Figure~\ref{fig:threatmodel}. We define two key parties involved: \textbf{(1) Image Owner} and \textbf{(2) Image User}. The specific objectives of each party are outlined as follows:

\noindent\textbf{Image Owner:} Image Owner holds the copyright for an image dataset that may be utilized by Image User to fine-tune a diffusion model. For the purpose of copyright protection, Image Owner aims to ensure that the output of the fine-tuned model remain copyright information. Furthermore, in scenarios where the generated content is considered inappropriate, traceability to the training dataset utilized during fine-tuning should be implemented to support accountability. Therefore, Image Owner employs dataset watermarking techniques to achieve copyright protection and traceability. The embedded watermarks should satisfy the imperceptibility requirement, ensuring that it remains undetectable to human eyes and does not interfere with the generative performance of the fine-tuned diffusion model.

\noindent\textbf{Image User:} Image User collects multiple images related to the same character or style and subsequently fine-tune a diffusion model using this dataset. In cases where the fine-tuning dataset is protected through dataset watermarking, the outputs generated by the fine-tuned model should retain the embedded watermark information. On the Image User side, dataset watermarking faces three primary challenges: (1) Image User may apply various fine-tuning methods to address diverse generation tasks, with both the specific fine-tuning methods and tasks being unknown to Image Owner; (2) Image User may employ a mixed dataset containing both watermarked and original images for fine-tuning; and (3) malicious Image User may apply post-processing techniques to remove dataset watermarks prior to fine-tuning in an attempt to infringe copyrights. Therefore, dataset watermarking techniques must exhibit effectiveness across all three aforementioned challenging scenarios.
\subsection{Dimensions of Evaluation}
Building upon the above analysis of the threat model, we categorize the requirements of dataset watermarking in diffusion models into three key dimensions, which collectively constitute the evaluation framework: \textbf{(1) Universality:} the watermarking methods should be adaptive to various fine-tuning approaches and diverse generation tasks; \textbf{(2) Transmissibility:} the watermarking methods should be capable of preserving and propagating the watermarking effect, even when only a portion of the images in the entire fine-tuning dataset are watermarked; \textbf{(3) Robustness:} the watermarking methods should be resilience against post-processing operations including both common image quality degradation and tailored watermark removal attack. The following sections will present a comprehensive evaluation of existing dataset watermarking techniques for diffusion model, based on these three dimensions.

\section{Comprehensive Benchmark}
\label{sec:benchmark}
This section presents a comprehensive benchmark that is established for dataset watermarking techniques within the context of tracking diffusion model fine-tuning. 
\subsection{Experimental Settings}
We evaluate four state-of-the-art dataset watermarking methods which are open source: DIAGNOSIS~\cite{wang2024diagnosis}, DiffusionShield~\cite{cui2025diffusionshield}, SIREN~\cite{li2025towards}, and WatermarkDM~\cite{zhao2023recipe}. In experiments, all fine-tuning methods are based on stable diffusion 1.4 (SD1.4)\cite{rombach2022high}. The fine-tuning steps were 50, and the rest followed the settings in the original paper. To evaluate the universality across various generation tasks, we select three datasets in distinct styles for fine-tuning: CelebA-HQ~\cite{liu2015deep}, Pokemon~\cite{pinkney2022pokemon}, and WikiArt~\cite{wikiart}. To quantify the performance of these methods, we employ three evaluation metrics: FID \cite{heusel2017gans}, CLIP similarity \cite{wang2023exploring}, and watermarking detection accuracy (referred to as Acc in the tables). All experiments are conducted on 4 A800 GPUs.
\begin{table}[!t]
\renewcommand\arraystretch{0.6}
\centering
\caption{Specific configurations for the four fine-tuning methods. In UNet, "CA" denotes the Cross-Attention, and "TE" (te) refers to the Text Encoder. Among the four fine-tuning approaches, the text encoder can be configured in either a trainable mode (w/ te) or a frozen mode (w/o te).
}
\begin{adjustbox}{width=\linewidth}
\begin{tabular}{ccccc c}
\toprule
\multicolumn{2}{c}{\multirow{2}{*}{\textbf{Fine-Tuning Method}}} & \multicolumn{4}{c}{\textbf{Trainable Layers}} \\ \cmidrule{3-6}
\multicolumn{2}{c}{} & FULL-UNet & VAE & CA & TE \\
\midrule
\multirow{2}{*}{Text-to-Image} & w te & \ding{51} & \ding{51} & \ding{55} & \ding{51} \\
& w/o te & \ding{51} & \ding{55} & \ding{51} & \ding{55} \\
\midrule
\multirow{2}{*}{LoRA} & w te & \ding{55} & \ding{55} & \ding{51} & \ding{51} \\ 
& w/o te & \ding{55} & \ding{55} & \ding{51} & \ding{55} \\
\midrule
\multirow{2}{*}{DreamBooth} & w te & \ding{51} & \ding{55} & \ding{51} & \ding{51} \\
& w/o te & \ding{51} & \ding{55} & \ding{51} & \ding{55} \\
\midrule
\multirow{2}{*}{Textual Inversion} & w te & \ding{55} & \ding{55} & \ding{55} & \ding{51} \\
& w/o te & \ding{55} & \ding{55} & \ding{55} & \ding{55} \\
\bottomrule
\end{tabular}
\end{adjustbox}
\label{tab_fineconfig}
\end{table}
\begin{table*}[!t]
\setlength\tabcolsep{1pt}
\renewcommand\arraystretch{1.0}
\centering
\caption{The results of different watermark protection methods using various fine-tuning methods on CelebA-HQ, Pokémon and WikiArt datasets. The reference, best, and worst performance are marked by \textbf{bold}, \textcolor{red}{red}, and \textcolor{blue}{blue}, respectively.}
\begin{adjustbox}{width=\textwidth}
\begin{tabular}{c c|c ccc|ccc|ccc|ccc}
\toprule
\multirow{2.5}{*}{\textbf{Dataset}} & \multirow{2.5}{*}{\textbf{te}} & \textbf{FT} & \multicolumn{3}{c|}{\textbf{Text-to-Image}} & \multicolumn{3}{c|}{\textbf{LoRA}} & \multicolumn{3}{c|}{\textbf{DreamBooth}} & \multicolumn{3}{c}{\textbf{Textual Inversion}} \\
\cmidrule(lr){3-15}
&  & \textbf{Metrics} & \textbf{CLIP-T$\uparrow$} & \textbf{FID$\downarrow$} & \textbf{Acc.(\%)$\uparrow$} & \textbf{CLIP-T$\uparrow$} & \textbf{FID$\downarrow$} & \textbf{Acc.(\%)$\uparrow$} & \textbf{CLIP-T$\uparrow$} & \textbf{FID$\downarrow$} & \textbf{Acc.(\%)$\uparrow$} & \textbf{CLIP-T$\uparrow$} & \textbf{FID$\downarrow$} & \textbf{Acc.(\%)$\uparrow$} \\
\cmidrule(lr){1-15}
\multirow{10.5}{*}{\textbf{CelebA-HQ}} & \multirow{5}{*}{\textbf{w/o(without)}} & Clean & \textbf{0.2309} & \textbf{224.70} & \textbf{N/A} & \textbf{0.2625} & \textbf{221.10} & \textbf{N/A} & \textbf{0.2565} & \textbf{207.94} & \textbf{N/A} & \textbf{0.2630} & \textbf{222.49} & \textbf{N/A} \\
& & DIAGNOSIS & 0.1947 & \textcolor{red}{247.45} & 64.00 & \textcolor{blue}{0.2038} & \textcolor{red}{226.11} & \textcolor{blue}{40.00} & \textcolor{blue}{0.2384} & \textcolor{blue}{261.38} & \textcolor{blue}{20.00} & \textcolor{blue}{0.2304} & 280.24 & \textcolor{blue}{36.00} \\
& & DiffusionShield & \textcolor{red}{0.2184} & 259.99 & \textcolor{red}{99.83} & \textcolor{red}{0.2603} & \textcolor{blue}{276.64} & \textcolor{red}{99.00} & 0.2511 & \textcolor{red}{245.11} & \textcolor{red}{100.00} & 0.2390 & \textcolor{red}{264.55} & \textcolor{red}{98.78} \\
& & WatermarkDM & \textcolor{blue}{0.1814} & \textcolor{blue}{286.58} & 98.44 & 0.2328 & 239.68 & 96.88 & \textcolor{red}{0.2600} & 246.40 & 95.31 & 0.2611 & \textcolor{blue}{285.06} & 96.88 \\
& & SIREN & 0.2155 & 255.96 & \textcolor{blue}{55.25} & 0.2593 & 274.40 & 55.04 & 0.2552 & 260.58 & 55.83 & \textcolor{red}{0.2636} & 270.83 & 53.83 \\
\cmidrule(lr){2-15}
& \multirow{5}{*}{\textbf{w(with)}} & Clean & \textbf{0.2323} & \textbf{213.85} & \textbf{N/A} & \textbf{0.2459} & \textbf{264.44} & \textbf{N/A} & \textbf{0.2684} & \textbf{229.83} & \textbf{N/A} & \textbf{0.2649} & \textbf{226.88} & \textbf{N/A} \\
& & DIAGNOSIS & 0.1946 & 253.52 & 92.00 & \textcolor{blue}{0.2188} & 274.43 & \textcolor{blue}{46.00} & \textcolor{blue}{0.2634} & 257.45 & 84.00 & \textcolor{blue}{0.2311} & 274.67 & \textcolor{blue}{50.00} \\
& & DiffusionShield & 0.2023 & \textcolor{red}{251.19} & \textcolor{red}{100.00} & \textcolor{red}{0.2612} & \textcolor{blue}{283.58} & \textcolor{red}{100.00} & 0.2647 & 264.84 & \textcolor{red}{100.00} & 0.2358 & \textcolor{red}{271.05} & \textcolor{red}{100.00} \\
& & WatermarkDM & \textcolor{blue}{0.1819} & \textcolor{blue}{307.61} & 95.31 & 0.2350 & \textcolor{red}{239.29} & 93.75 & 0.2692 & \textcolor{red}{252.46} & 90.63 & 0.2600 & \textcolor{blue}{280.33} & 92.19 \\
& & SIREN & \textcolor{red}{0.2174} & 256.78 & \textcolor{blue}{54.63} & 0.2576 & 270.40 & 54.13 & \textcolor{red}{0.2742} & \textcolor{blue}{271.23} & \textcolor{blue}{54.58} & \textcolor{red}{0.2629} & 274.55 & 54.04 \\
\midrule
\multirow{10.5}{*}{\textbf{Pokémon}} & \multirow{5}{*}{\textbf{w/o(without)}} & Clean & \textbf{0.2281} & \textbf{127.07} & \textbf{N/A} & \textbf{0.2856} & \textbf{158.83} & \textbf{N/A} & \textbf{0.2931} & \textbf{218.68} & \textbf{N/A} & \textbf{0.2864} & \textbf{184.81} & \textbf{N/A} \\
& & DIAGNOSIS & \textcolor{red}{0.2171} & \textcolor{blue}{303.99} & 84.00 & 0.2597 & \textcolor{blue}{277.88} & 94.00 & \textcolor{blue}{0.2922} & \textcolor{blue}{273.28} & 70.00 & 0.2860 & \textcolor{red}{259.04} & 70.00 \\
& & DiffusionShield & \textcolor{blue}{0.2013} & 206.85 & \textcolor{red}{96.19} & 0.2843 & 247.60 & \textcolor{red}{98.78} & 0.2951 & 223.50 & \textcolor{red}{98.34} & \textcolor{blue}{0.2825} & \textcolor{blue}{267.79} & \textcolor{red}{98.81} \\
& & WatermarkDM & 0.2050 & 258.19 & \textcolor{red}{100.00} & \textcolor{blue}{0.2447} & \textcolor{red}{191.65} & 98.44 & 0.2932 & 231.38 & 98.44 & \textcolor{red}{0.2924} & 260.91 & 95.31 \\
& & SIREN & 0.2098 & \textcolor{red}{201.16} & \textcolor{blue}{57.04} & \textcolor{red}{0.2926} & 235.50 & \textcolor{blue}{57.04} & \textcolor{red}{0.2953} & \textcolor{red}{211.87} & \textcolor{blue}{57.08} & 0.2907 & 267.77 & \textcolor{blue}{55.96} \\
\cmidrule(lr){2-15}
& \multirow{5}{*}{\textbf{w(with)}} & Clean & \textbf{0.2060} & \textbf{136.15} & \textbf{N/A} & \textbf{0.2592} & \textbf{212.30} & \textbf{N/A} & \textbf{0.2902} & \textbf{200.60} & \textbf{N/A} & \textbf{0.2859} & \textbf{179.20} & \textbf{N/A} \\
& & DIAGNOSIS & 0.2050 & \textcolor{blue}{338.90} & 98.00 & 0.2597 & \textcolor{blue}{262.30} & 96.00 & 0.2916 & \textcolor{blue}{270.58} & \textcolor{blue}{18.00} & \textcolor{blue}{0.2823} & 264.96 & \textcolor{blue}{44.00} \\
& & DiffusionShield & \textcolor{red}{0.2139} & 220.54 & \textcolor{red}{99.64} & \textcolor{red}{0.2899} & 234.32 & \textcolor{red}{98.92} & \textcolor{red}{0.2929} & \textcolor{red}{213.00} & \textcolor{red}{99.91} & 0.2844 & 268.66 & \textcolor{red}{99.87} \\
& & WatermarkDM & \textcolor{blue}{0.1943} & 273.59 & 96.88 & \textcolor{blue}{0.2283} & \textcolor{red}{204.78} & 98.44 & 0.2913 & 230.89 & 90.63 & \textcolor{red}{0.2938} & \textcolor{blue}{273.69} & 87.50 \\
& & SIREN & 0.2131 & \textcolor{red}{205.86} & \textcolor{blue}{56.88} & 0.2839 & 244.35 & \textcolor{blue}{56.58} & \textcolor{blue}{0.2868} & 223.12 & 57.75 & 0.2864 & \textcolor{red}{247.05} & 56.00 \\
\midrule
\multirow{10.5}{*}{\textbf{WikiArt}} & \multirow{5}{*}{\textbf{w/o(without)}} & Clean & \textbf{0.1401} & \textbf{320.86} & \textbf{N/A} & \textbf{0.2703} & \textbf{312.60} & \textbf{N/A} & \textbf{0.2670} & \textbf{320.12} & \textbf{N/A} & \textbf{0.2735} & \textbf{320.63} & \textbf{N/A} \\
& & DIAGNOSIS & 0.1458 & \textcolor{blue}{365.44} & 90.00 & 0.2678 & 308.55 & 82.00 & 0.2680 & \textcolor{red}{310.55} & \textcolor{blue}{30.00} & \textcolor{blue}{0.2765} & 318.46 & 66.00 \\
& & DiffusionShield & \textcolor{red}{0.1671} & \textcolor{red}{326.28} & \textcolor{red}{100.00} & \textcolor{red}{0.2742} & \textcolor{red}{307.60} & \textcolor{red}{97.73} & \textcolor{red}{0.2755} & 318.46 & \textcolor{red}{99.91} & 0.2770 & \textcolor{red}{312.09} & \textcolor{red}{98.77} \\
& & WatermarkDM & \textcolor{blue}{0.1388} & 362.88 & 90.56 & \textcolor{blue}{0.1624} & \textcolor{blue}{352.12} & 71.34 & \textcolor{blue}{0.2480} & \textcolor{blue}{336.02} & 51.31 & \textcolor{red}{0.2791} & 324.83 & 91.47 \\
& & SIREN & 0.1574 & 329.96 & \textcolor{blue}{52.08} & 0.2713 & 325.00 & \textcolor{blue}{53.46} & 0.2662 & 319.62 & 53.54 & 0.2789 & \textcolor{blue}{329.95} & \textcolor{blue}{54.54} \\
\cmidrule(lr){2-15}
& \multirow{5}{*}{\textbf{w(with)}} & Clean & \textbf{0.1482} & \textbf{317.17} & \textbf{N/A} & \textbf{0.2703} & \textbf{306.54} & \textbf{N/A} & \textbf{0.2720} & \textbf{325.00} & \textbf{N/A} & \textbf{0.2745} & \textbf{315.48} & \textbf{N/A} \\
& & DIAGNOSIS & \textcolor{blue}{0.1164} & \textcolor{blue}{384.54} & \textcolor{blue}{16.00} & 0.2612 & \textcolor{red}{301.19} & \textcolor{blue}{44.00} & 0.2690 & \textcolor{red}{316.08} & 74.00 & 0.2785 & 324.58 & \textcolor{blue}{30.00} \\
& & DiffusionShield & \textcolor{red}{0.1782} & \textcolor{red}{294.04} & \textcolor{red}{100.00} & 0.2657 & 306.48 & \textcolor{red}{99.80} & \textcolor{red}{0.2751} & 319.39 & \textcolor{red}{97.41} & 0.2787 & 326.07 & \textcolor{red}{99.77} \\
& & WatermarkDM & 0.1410 & 361.31 & 73.94 & \textcolor{blue}{0.1659} & \textcolor{blue}{351.42} & 68.00 & \textcolor{blue}{0.2606} & \textcolor{blue}{333.59} & \textcolor{blue}{51.47} & \textcolor{blue}{0.2777} & \textcolor{blue}{326.14} & 49.84 \\
& & SIREN & 0.1528 & 306.82 & 51.63 & \textcolor{red}{0.2704} & 320.81 & 53.63 & 0.2742 & 329.37 & 53.71 & \textcolor{red}{0.2821} & \textcolor{red}{321.96} & 53.67 \\
\bottomrule
\end{tabular}
\end{adjustbox}
\label{tab_finetuning}
\end{table*}

\subsection{Universality Evaluation}
To evaluate the cross-method universality, we conduct assessments of these dataset watermarking techniques under various fine-tuning methodologies. Specifically, four fine-tuning methods with distinct configurations are utilized in the experiments, as detailed in Table~\ref{tab_fineconfig}. The table summarizes the trainable modules associated with each fine-tuning approach, where "te" denotes whether the text encoder is frozen during the fine-tuning process. All methods employ the default hyperparameter configurations specified in original papers, ensuring convergence of the fine-tuning process to an optimal state. Quantitative results obtained from evaluations on three datasets using four fine-tuning approaches are presented in Table~\ref{tab_finetuning}. Regarding the generation quality after fine-tuning, fine-tuning with watermarked images adversely affects performance, potentially leading to a decrease in FID score and an increase in CLIP-T score. Specifically, the use of watermarked images for fine-tuning resulted in the most significant decline in generation performance on the Pokémon dataset. 

To ensure a fair comparison of each watermarking method's ability to detect watermarks from generated images, we calibrate the watermark embedding strength across all methods, thereby guaranteeing that the fine-tuned generation results based on each watermarked dataset exhibit comparable performance in terms of FID and CLIP-T metrics. The detection accuracy results of watermark extraction are presented in Table~\ref{tab_finetuning}. The experimental results demonstrate that DiffusionShield achieves the best performance, exhibiting a watermark detection accuracy approaching 100\% across various datasets and fine-tuning methods. Conversely, SIREN exhibits the lowest performance, with a detection accuracy of approximately 50\% across all datasets and fine-tuning methods, which is equivalent to random prediction. Consequently, SIREN struggles to effectively trace the origin data of generated images, such as portraits and artworks, in real-world applications. The remaining two methods exhibit comparatively favorable performance under specific experimental conditions. Specifically, WatermarkDM attains a watermark detection accuracy exceeding 90\% on the WikiArt dataset for both Text-to-Image and Textual Inversion fine-tuning, suggesting a degree of adaptability of the watermark to varying text conditions. Nevertheless, the detection accuracy of this method declines substantially when fine-tuning is applied to the text encoder. DIAGNOSIS presents satisfactory performance across four fine-tuning methods on the Pokémon dataset. The adoption of fine-tuning the text encoder also significantly decreases the detection accuracy.
\begin{table*}[!t]
\setlength\tabcolsep{1pt}
\renewcommand\arraystretch{1.0}
\centering
\caption{Summary of watermark protection ratio results using different fine-tuning methods on the CelebA-HQ dataset. The best and worst performance are marked by \textcolor{red}{red}, and \textcolor{blue}{blue}, respectively.}
\begin{adjustbox}{width=\textwidth}
\begin{tabular}{c|c|c ccc|ccc|ccc|ccc}
\toprule   
\multirow{2.5}{*}{\textbf{\makecell{Protection\\Ratio}}} & \multirow{2.5}{*}{\textbf{te}} & \textbf{FT} & \multicolumn{3}{c|}{\textbf{Text-to-Image}} & \multicolumn{3}{c|}{\textbf{LoRA}} & \multicolumn{3}{c|}{\textbf{DreamBooth}} & \multicolumn{3}{c}{\textbf{Textual Inversion}} \\
\cmidrule(lr){3-15}
&  & \textbf{Metrics} & \textbf{CLIP-T$\uparrow$} & \textbf{FID$\downarrow$} & \textbf{Acc.(\%)$\uparrow$} & \textbf{CLIP-T$\uparrow$} & \textbf{FID$\downarrow$} & \textbf{Acc.(\%)$\uparrow$} & \textbf{CLIP-T$\uparrow$} & \textbf{FID$\downarrow$} & \textbf{Acc.(\%)$\uparrow$} & \textbf{CLIP-T$\uparrow$} & \textbf{FID$\downarrow$} & \textbf{Acc.(\%)$\uparrow$} \\
\cmidrule(lr){1-15}
\multirow{8.5}{*}{\textbf{20\%}} & \multirow{4}{*}{\textbf{w/o(without)}} & DIAGNOSIS & \textcolor{blue}{0.2047} & \textcolor{blue}{259.62} & 80.00 & 0.2601 & 279.06 & \textcolor{blue}{40.00} & \textcolor{red}{0.2632} & 255.46 & \textcolor{blue}{16.00} & 0.2634 & 283.63 & \textcolor{blue}{44.00} \\
& & DiffusionShield & 0.2049 & \textcolor{red}{252.40} & \textcolor{red}{100.00} & \textcolor{blue}{0.2585} & 278.47 & \textcolor{red}{100.00} & 0.2592 & 252.05 & \textcolor{red}{99.56} & 0.2672 & \textcolor{blue}{289.95} & \textcolor{red}{99.91} \\
& & WatermarkDM & \textcolor{red}{0.2191} & 257.56 & 98.44 & 0.2606 & \textcolor{red}{276.72} & 89.06 & \textcolor{blue}{0.2540} & \textcolor{red}{249.44} & 85.94 & \textcolor{blue}{0.2618} & 285.29 & 92.19 \\
& & SIREN & 0.2145 & 256.13 & \textcolor{blue}{57.25} & \textcolor{red}{0.2663} & \textcolor{blue}{283.66} & 53.92 & 0.2607 & \textcolor{blue}{272.31} & 55.13 & \textcolor{red}{0.2668} & \textcolor{red}{280.46} & 55.17 \\
\cmidrule(lr){2-15}
& \multirow{4}{*}{\textbf{w(with)}} & DIAGNOSIS & \textcolor{red}{0.2119} & \textcolor{red}{251.84} & 76.00 & 0.2615 & 279.03 & \textcolor{blue}{26.00} & \textcolor{blue}{0.2630} & 265.31 & \textcolor{blue}{46.00} & \textcolor{red}{0.2674} & 275.48 & \textcolor{blue}{36.00} \\
& & DiffusionShield & \textcolor{blue}{0.1987} & 257.69 & \textcolor{red}{100.00} & 0.2586 & \textcolor{red}{273.14} & \textcolor{red}{98.88} & 0.2697 & \textcolor{blue}{277.35} & \textcolor{red}{100.00} & \textcolor{blue}{0.2628}& 277.96 & \textcolor{red}{100.00} \\
& & WatermarkDM & 0.2120 & \textcolor{blue}{268.10} & 96.88 & \textcolor{blue}{0.2530} & 282.16 & 95.31 & 0.2721 & \textcolor{red}{260.63} & 95.31 & 0.2661 & \textcolor{red}{267.53} & 93.75 \\
& & SIREN & 0.2025 & 261.54 & 57.58 & \textcolor{red}{0.2701} & \textcolor{blue}{284.60} & 53.67 & \textcolor{red}{0.2737} & 272.81 & 56.13 & 0.2634 & \textcolor{blue}{279.70} & 54.21 \\
\midrule
\multirow{8.5}{*}{\textbf{40\%}} & \multirow{4}{*}{\textbf{w/o(without)}} & DIAGNOSIS & \textcolor{red}{0.2025} & 261.50 & \textcolor{blue}{44.00} & \textcolor{red}{0.2683} & \textcolor{blue}{287.50} & \textcolor{blue}{36.00} & 0.2598 & \textcolor{red}{247.40} & \textcolor{blue}{6.00} & 0.2605 & 275.94 & \textcolor{blue}{40.00} \\
& & DiffusionShield & 0.2048 & \textcolor{blue}{256.56} & \textcolor{red}{100.00} & \textcolor{blue}{0.2565} & \textcolor{red}{276.34} &\textcolor{red}{99.00} & 0.2534 & 259.52 & \textcolor{red}{99.00} & \textcolor{blue}{0.2600} & \textcolor{red}{271.86} & \textcolor{red}{100.00} \\
& & WatermarkDM & \textcolor{blue}{0.1870} & \textcolor{red}{253.48} & 89.06 & 0.2602 & 279.92 & 78.13 & \textcolor{blue}{0.2532} & 252.08 & 98.44 & 0.2657 & 278.50 & 90.63 \\
& & SIREN & 0.2024 & \textcolor{blue}{264.39} & 56.58 & 0.2658 & 281.70 & 54.29 & \textcolor{red}{0.2613} & \textcolor{blue}{262.93} & 54.04 & \textcolor{red}{0.2671} & \textcolor{blue}{279.16} & 54.00 \\
\cmidrule(lr){2-15}
& \multirow{4}{*}{\textbf{w(with)}} & DIAGNOSIS & \textcolor{red}{0.2101} & \textcolor{red}{247.40} & 80.00 & 0.2646 & 275.39 & \textcolor{blue}{36.00} & \textcolor{blue}{0.2643} & 274.05 & \textcolor{blue}{52.00} & 0.2616 & 272.49 & 78.00 \\
& & DiffusionShield & 0.2085 & 257.46 & \textcolor{red}{100.00} & 0.2645 & 276.76 & \textcolor{red}{100.00} & 0.2682 & 266.83 & \textcolor{red}{100.00} & \textcolor{red}{0.2644} & \textcolor{blue}{275.95} & \textcolor{red}{98.95} \\
& & WatermarkDM & \textcolor{blue}{0.1903} & 255.94 & 98.44 & \textcolor{blue}{0.2570} & \textcolor{blue}{278.65} & 85.94 & \textcolor{red}{0.2729} & \textcolor{red}{265.11} & 96.88 & \textcolor{blue}{0.2582} & 275.12 & 95.31 \\
& & SIREN & 0.1966 & \textcolor{blue}{262.58} & \textcolor{blue}{55.29} & \textcolor{red}{0.2712} & \textcolor{red}{275.09} & 55.29 & 0.2692 & \textcolor{blue}{281.19} & 55.25 & 0.2611 & \textcolor{red}{268.52} & \textcolor{blue}{54.58} \\
\midrule
\multirow{8.5}{*}{\textbf{60\%}} & \multirow{4}{*}{\textbf{w/o(without)}} & DIAGNOSIS & \textcolor{red}{0.2139} & 250.92 & \textcolor{blue}{20.00} & 0.2599 & \textcolor{red}{273.85} & 92.00 & 0.2628 & 250.70 & \textcolor{blue}{18.00} & 0.2646 & 273.28 & 52.00 \\
& & DiffusionShield & 0.2081 & \textcolor{red}{232.99} & \textcolor{red}{100.00} & \textcolor{blue}{0.2573} & 280.14 & \textcolor{red}{99.73} & 0.2517 & \textcolor{red}{246.38} & \textcolor{red}{99.91} & 0.2642 & 269.59 & \textcolor{red}{99.19}\\
& & WatermarkDM & \textcolor{blue}{0.2019} & \textcolor{blue}{259.35} & 95.31 & 0.2613 & 274.82 & 92.19 & \textcolor{blue}{0.2511} & 250.54 & 90.63 & \textcolor{blue}{0.2590} & \textcolor{blue}{288.34} & 89.06 \\
& & SIREN & 0.2078 & 256.26 & 53.25 & \textcolor{red}{0.2617} & \textcolor{blue}{281.14} & 54.67 & \textcolor{red}{0.2673} & \textcolor{blue}{257.28} & 56.33 & \textcolor{red}{0.2648} & \textcolor{red}{268.78} & 54.79 \\
\cmidrule(lr){2-15}
& \multirow{4}{*}{\textbf{w(with)}} & DIAGNOSIS & \textcolor{red}{0.2066} & 246.85 & \textcolor{blue}{34.00} & 0.2556 & \textcolor{red}{273.47} & 70.00 & \textcolor{blue}{0.2631} & 268.76 & \textcolor{blue}{26.00} & 0.2654 & \textcolor{red}{269.74} & \textcolor{blue}{20.00} \\
& & DiffusionShield & 0.2064 & \textcolor{red}{241.85} & \textcolor{red}{99.94} & \textcolor{blue}{0.2531} & 278.91 & \textcolor{red}{98.78} & 0.2638 & 262.69 & \textcolor{red}{100.00} & \textcolor{blue}{0.2630} & \textcolor{blue}{280.64} & \textcolor{red}{100.00} \\
& & WatermarkDM & \textcolor{blue}{0.2002} & \textcolor{blue}{269.80} & 93.75 & 0.2581 & 279.70 & 92.19 & 0.2699 & \textcolor{red}{261.96} & 93.75 & 0.2643 & 275.62 & 90.63 \\
& & SIREN & 0.2033 & 268.06 & 53.88 & \textcolor{red}{0.2587} & \textcolor{blue}{280.42} & 54.33 & \textcolor{red}{0.2714} & \textcolor{blue}{280.42} & 55.42 & \textcolor{red}{0.2664} & 271.40 & 54.67 \\
\midrule
\multirow{8.5}{*}{\textbf{80\%}} & \multirow{4}{*}{\textbf{w/o(without)}} & DIAGNOSIS & \textcolor{blue}{0.1962} & \textcolor{blue}{268.29} & 60.00 & \textcolor{blue}{0.2528} & 280.73 & \textcolor{blue}{50.00} & \textcolor{red}{0.2667} & \textcolor{red}{246.12} & 70.00 & \textcolor{red}{0.2647} & 275.99 & \textcolor{blue}{50.00} \\
& & DiffusionShield & \textcolor{red}{0.2126} & 251.40 & \textcolor{red}{100.00} & 0.2543 & 276.66 & \textcolor{red}{99.97} & \textcolor{blue}{0.2513} & 249.16 & \textcolor{red}{99.19} & \textcolor{blue}{0.2632} & \textcolor{blue}{279.37} & \textcolor{red}{100.00} \\
& & WatermarkDM & 0.1969 & 257.88 & 98.44 & 0.2577 & \textcolor{red}{273.84} & 100.00 & 0.2594 & 246.49 & 98.99 & 0.2637 & \textcolor{red}{272.91} & 98.44 \\
& & SIREN & 0.2024 & \textcolor{red}{248.74} & \textcolor{blue}{54.63} & \textcolor{red}{0.2696} & \textcolor{blue}{282.74} & 54.13 & 0.2598 & \textcolor{blue}{255.20} & \textcolor{blue}{56.67} & \textcolor{blue}{0.2632} & 276.49 & 54.08 \\
\cmidrule(lr){2-15}
& \multirow{4}{*}{\textbf{w(with)}} & DIAGNOSIS & 0.1967 & \textcolor{blue}{268.98} & \textcolor{blue}{52.00} & 0.2581 & 281.58 & 70.00 & 0.2657 & \textcolor{blue}{269.38} & 60.00 & \textcolor{blue}{0.2600} & 280.34 & \textcolor{blue}{50.00} \\
& & DiffusionShield & 0.1971 & 265.04 & \textcolor{red}{99.80} & \textcolor{blue}{0.2565} & \textcolor{blue}{286.48} & \textcolor{red}{100.00} & 0.2649 & 264.26 & \textcolor{red}{100.00} & 0.2626 & \textcolor{blue}{280.84} & \textcolor{red}{99.00} \\
& & WatermarkDM & \textcolor{blue}{0.1926} & 266.00 & 98.44 & 0.2664 & 285.49 & 98.44 & \textcolor{blue}{0.2625} & \textcolor{red}{260.70} & 96.88 & \textcolor{red}{0.2696} & \textcolor{red}{267.71} & 95.31 \\
& & SIREN & \textcolor{red}{0.1996} & \textcolor{red}{264.95} & 54.13 & \textcolor{red}{0.2709} & \textcolor{red}{278.61} & \textcolor{blue}{54.08} & \textcolor{red}{0.2727} & 267.00 & \textcolor{blue}{55.96} & 0.2662 & 272.22 & 54.67 \\
\bottomrule
\end{tabular}
\end{adjustbox}
\label{tab_prot_ratio}
\end{table*}
\subsection{Transmissibility Evaluation}
In real-world applications, users may employ mixed datasets comprising both unwatermarked original images and watermarked traceable images to fine-tune diffusion models. Therefore, it is essential to assess the transmissibility performance of existing dataset watermarks, specifically whether the watermarks remain intact after the model has been fine-tuned using a subset of watermarked images. To this end, we conduct mixed fine-tuning using both original and watermarked data. Specifically, the proportion of watermarked images is set at 20\%, 40\%, 60\%, and 80\% respectively. Compared to the results obtained when fine-tuning with fully watermarked images in Table~\ref{tab_prot_ratio}, the detection accuracy of DIAGNOSIS decreases significantly, whereas the DiffusionShield remains largely unaffected. The other two methods fail to demonstrate the traceability of the dataset watermark to the results generated after fine-tuning.
\begin{table*}[!tb]
\setlength\tabcolsep{1pt}
\renewcommand\arraystretch{1.0}
\centering
\caption{Summary of natural distortion to watermark protection results under different fine-tuning methods on CelebA-HQ. The best and worst performance are marked by \textcolor{red}{red}, and \textcolor{blue}{blue}, respectively.}
\begin{adjustbox}{width=\textwidth}
\begin{tabular}{c c|c ccc|ccc|ccc|ccc}
\toprule
\multirow{2.5}{*}{\textbf{\makecell{Distortion\\Type}}} & \multirow{2.5}{*}{\textbf{te}} & \textbf{FT} & \multicolumn{3}{c|}{\textbf{Text-to-Image}} & \multicolumn{3}{c|}{\textbf{LoRA}} & \multicolumn{3}{c|}{\textbf{DreamBooth}} & \multicolumn{3}{c}{\textbf{Textual Inversion}} \\
\cmidrule(lr){3-15}
&  & \textbf{Metrics} & \textbf{CLIP-T$\uparrow$} & \textbf{FID$\downarrow$} & \textbf{Acc.(\%)$\uparrow$} & \textbf{CLIP-T$\uparrow$} & \textbf{FID$\downarrow$} & \textbf{Acc.(\%)$\uparrow$} & \textbf{CLIP-T$\uparrow$} & \textbf{FID$\downarrow$} & \textbf{Acc.(\%)$\uparrow$} & \textbf{CLIP-T$\uparrow$} & \textbf{FID$\downarrow$} & \textbf{Acc.(\%)$\uparrow$} \\
\cmidrule(lr){1-15}
\multirow{8.5}{*}{\textbf{Blur}} & \multirow{4}{*}{\textbf{w/o(without)}} & DIAGNOSIS & 0.2136 & \textcolor{red}{239.78} & \textcolor{red}{100.00} & \textcolor{blue}{0.2547} & 286.35 & 64.00 & 0.2503 & 262.86 & 68.00 & \textcolor{blue}{0.2604} & \textcolor{red}{267.97} & \textcolor{blue}{26.00} \\
& & DiffusionShield & 0.2067 & \textcolor{blue}{386.99} & \textcolor{red}{100.00} & 0.2582 & \textcolor{blue}{289.61} & \textcolor{red}{100.00} & \textcolor{blue}{0.2383} & \textcolor{blue}{351.66} & \textcolor{red}{100.00} & \textcolor{red}{0.2660} & 279.73 & \textcolor{red}{100.00} \\
& & WatermarkDM & \textcolor{blue}{0.2040} & 276.36 & 98.44 & 0.2588 & \textcolor{red}{270.90} & 89.06 & \textcolor{red}{0.2523} & 276.02 & 95.31 & 0.2617 & 279.46 & 92.19 \\
& & SIREN & \textcolor{red}{0.2246} & 246.81 & \textcolor{blue}{51.71} & \textcolor{red}{0.2615} & 279.04 & \textcolor{blue}{52.17} & 0.2426 & \textcolor{red}{259.72} & \textcolor{blue}{52.33} & 0.2648 & \textcolor{blue}{282.85} & 51.88 \\
\cmidrule(lr){2-15}
& \multirow{4}{*}{\textbf{w(with)}} & DIAGNOSIS & 0.2073 & \textcolor{red}{238.28} & 92.00 & \textcolor{blue}{0.2526} & 278.83 & \textcolor{blue}{16.00}& 0.2504 & 271.76 & 60.00 & 0.2649 & 277.83 & 58.00 \\
& & DiffusionShield & 0.2070 & \textcolor{blue}{372.00} & \textcolor{red}{100.00} & 0.2538 & \textcolor{blue}{292.73} & \textcolor{red}{100.00} & \textcolor{blue}{0.2493} & \textcolor{blue}{321.52} & \textcolor{red}{100.00} & 0.2624 & \textcolor{blue}{278.38} & \textcolor{red}{100.00} \\
& & WatermarkDM & \textcolor{blue}{0.2053} & 273.67 & 98.44 & \textcolor{red}{0.2650} & \textcolor{red}{260.28} & 92.19 & \textcolor{red}{0.2607} & \textcolor{red}{261.10} & 93.75 & \textcolor{blue}{0.2622} & 271.23 & 90.63 \\
& & SIREN & \textcolor{red}{0.2149} & 245.44 & \textcolor{blue}{51.17} & 0.2564 & 273.18 & 52.08 & 0.2595 & 274.34 & \textcolor{blue}{51.63} & \textcolor{red}{0.2662} & \textcolor{red}{270.26} & \textcolor{blue}{52.04} \\
\midrule
\multirow{8.5}{*}{\textbf{JPEG}} & \multirow{4}{*}{\textbf{w/o(without)}} & DIAGNOSIS & 0.2097 & \textcolor{red}{253.33} & 54.00 & \textcolor{blue}{0.2513} & 278.11 & \textcolor{blue}{40.00} & \textcolor{blue}{0.2315} & \textcolor{blue}{267.46} & \textcolor{blue}{50.00} & 0.2615 & \textcolor{blue}{283.02} & 90.00 \\
& & DiffusionShield & \textcolor{red}{0.2223} & 254.74 & \textcolor{red}{100.00} & 0.2582 & \textcolor{blue}{279.15} & \textcolor{red}{100.00} & 0.2483 & \textcolor{red}{249.81} & \textcolor{red}{99.50} & \textcolor{blue}{0.2558} & 274.64 & \textcolor{red}{99.98} \\
& & WatermarkDM & \textcolor{blue}{0.2049} & \textcolor{blue}{284.84} & \textcolor{red}{100.00} & \textcolor{red}{0.2618} & \textcolor{red}{268.99} & 95.31 & 0.2440 & 262.14 & 98.44 & \textcolor{red}{0.2709} & \textcolor{red}{270.01} & 93.75 \\
& & SIREN & 0.2149 & 253.97 & \textcolor{blue}{52.58} & 0.2577 & 273.47 & 51.63 & \textcolor{red}{0.2591} & 254.38 & 54.42 & 0.2637 & 276.30 & \textcolor{blue}{51.88} \\
\cmidrule(lr){2-15}
& \multirow{4}{*}{\textbf{w(with)}} & DIAGNOSIS & \textcolor{blue}{0.2088} & 257.33 & 82.00 & \textcolor{blue}{0.2550} & \textcolor{red}{268.54} & 94.00 & \textcolor{blue}{0.2646} & \textcolor{blue}{270.64} & 76.00 & 0.2649 & 272.11 & \textcolor{red}{100.00} \\
& & DiffusionShield & \textcolor{red}{0.2194} & \textcolor{blue}{277.31} & \textcolor{red}{100.00} & \textcolor{red}{0.2648} & \textcolor{blue}{274.72} & \textcolor{red}{99.98} & 0.2703 & 259.33 & \textcolor{red}{99.67} & \textcolor{blue}{0.2598} & \textcolor{red}{270.88} & \textcolor{red}{100.00} \\
& & WatermarkDM & 0.2128 & 276.14 & 92.19 & 0.2624 & 270.72 & 95.31 & \textcolor{blue}{0.2646} & \textcolor{red}{253.28} & 90.63 & \textcolor{red}{0.2660} & 273.31 & 89.06 \\
& & SIREN & 0.2133 & \textcolor{red}{252.88} & \textcolor{blue}{53.25} & 0.2591 & 272.66 & \textcolor{blue}{52.21} & \textcolor{red}{0.2708} & 264.26 & \textcolor{blue}{53.08} & 0.2615 & \textcolor{blue}{287.28} & \textcolor{blue}{52.21} \\
\midrule
\multirow{8.5}{*}{\textbf{Noise}} & \multirow{4}{*}{\textbf{w/o(without)}} & DIAGNOSIS & \textcolor{red}{0.2213} & \textcolor{red}{264.12} & 60.00 & \textcolor{blue}{0.2614} & \textcolor{red}{279.73} & 70.00 & 0.2485 & \textcolor{red}{261.32} & 74.00 & 0.2694 & 272.99 & \textcolor{blue}{36.00} \\
& & DiffusionShield & \textcolor{blue}{0.1990} & \textcolor{blue}{388.39} & \textcolor{red}{82.16} & 0.2629 & \textcolor{blue}{307.26} & \textcolor{red}{92.52} & \textcolor{blue}{0.2400} & \textcolor{blue}{369.91} & \textcolor{red}{88.67} & \textcolor{red}{0.2697} & 274.24 & \textcolor{red}{100.00} \\
& & WatermarkDM & 0.2055 & 387.67 & 98.44 & 0.2640 & 283.23 & 98.44 & 0.2514 & 323.49 & 92.19 & \textcolor{blue}{0.2602} & \textcolor{red}{270.56} & 95.31 \\
& & SIREN & 0.2072 & 313.27 & \textcolor{blue}{50.67} & \textcolor{red}{0.2701} & 282.57 & \textcolor{blue}{51.25} & \textcolor{red}{0.2593} & 263.95 & \textcolor{blue}{49.96} & 0.2627 & \textcolor{blue}{279.44} & 52.38 \\
\cmidrule(lr){2-15}
& \multirow{4}{*}{\textbf{w(with)}} & DIAGNOSIS & \textcolor{red}{0.2157} & \textcolor{red}{270.54} & \textcolor{red}{100.00} & 0.2613 & 278.00 & \textcolor{blue}{34.00} & 0.2718 & 282.14 & \textcolor{red}{96.00} & \textcolor{red}{0.2678} & \textcolor{red}{267.09} & \textcolor{red}{100.00} \\
& & DiffusionShield & 0.2081 & 392.11 & 88.30 & \textcolor{blue}{0.2500} & \textcolor{blue}{303.06} & \textcolor{red}{93.08} & \textcolor{blue}{0.2490} & \textcolor{blue}{319.35} & 86.98 & \textcolor{blue}{0.2607} & \textcolor{blue}{275.41} & 98.89 \\
& & WatermarkDM & 0.2098 & \textcolor{blue}{422.80} & 96.88 & \textcolor{red}{0.2670} & \textcolor{red}{274.59} & 98.44 & \textcolor{red}{0.2805} & 287.30 & 90.63 & 0.2647 & 271.09 & 90.63 \\
& & SIREN & \textcolor{blue}{0.2015} & 341.88 & \textcolor{blue}{49.79} & 0.2576 & 283.16 & 50.54 & 0.2737 & \textcolor{red}{276.27} & \textcolor{blue}{50.08} & 0.2627 & 274.34 & \textcolor{blue}{52.17} \\
\bottomrule
\end{tabular}
\end{adjustbox}
\label{tab_distortion}
\end{table*}
\begin{figure*}
    \centering
    \includegraphics[width=\textwidth]{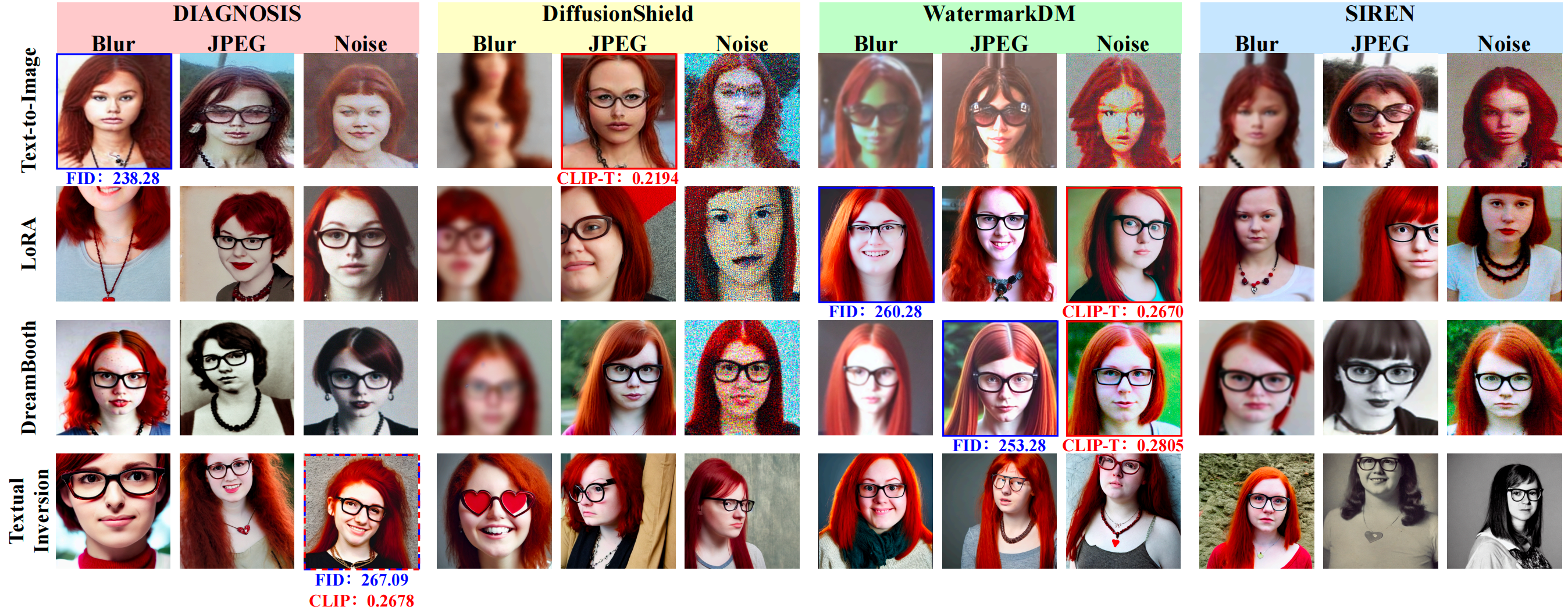}
    \caption{ The visualization of generation results after applying natural distortion. The figure indicates the optimal \textcolor{blue}{FID} score
 and \textcolor{red}{CLIP-T} similarity for each fine-tuning approach.}
    \label{fig:distortion_vis}
\end{figure*}
\subsection{Robustness Evaluation under Common Distortions}
Dataset watermarking methods may be vulnerable to potential post-processing operations, which can compromise the integrity of the embedded watermark signals. Hence, it is crucial to evaluate the robustness of existing dataset watermarking approaches against a variety of post-processing techniques. We categorize potential post-processing operations into two classes: common image processing techniques and specifically designed watermark removal methods. The first category of post-processing operations may lead to overall quality degradation in watermarked images, as shown in Figure~\ref{fig:distortion_vis}, consequently resulting in diminished performance during fine-tuning. We evaluate the robustness of existing methods under common image degradation operations, including Gaussian noise, Gaussian blur, and JPEG compression. These three types of image processing techniques are typically introduced during the transmission and processing of datasets, rather than being specifically applied for malicious attacks. As shown in Table~\ref{tab_distortion}, both DIAGNOSIS and DiffusionShield exhibit robustness in these conditions.
\begin{figure*}[!t]
  \centering
  \includegraphics[width=.98\textwidth]{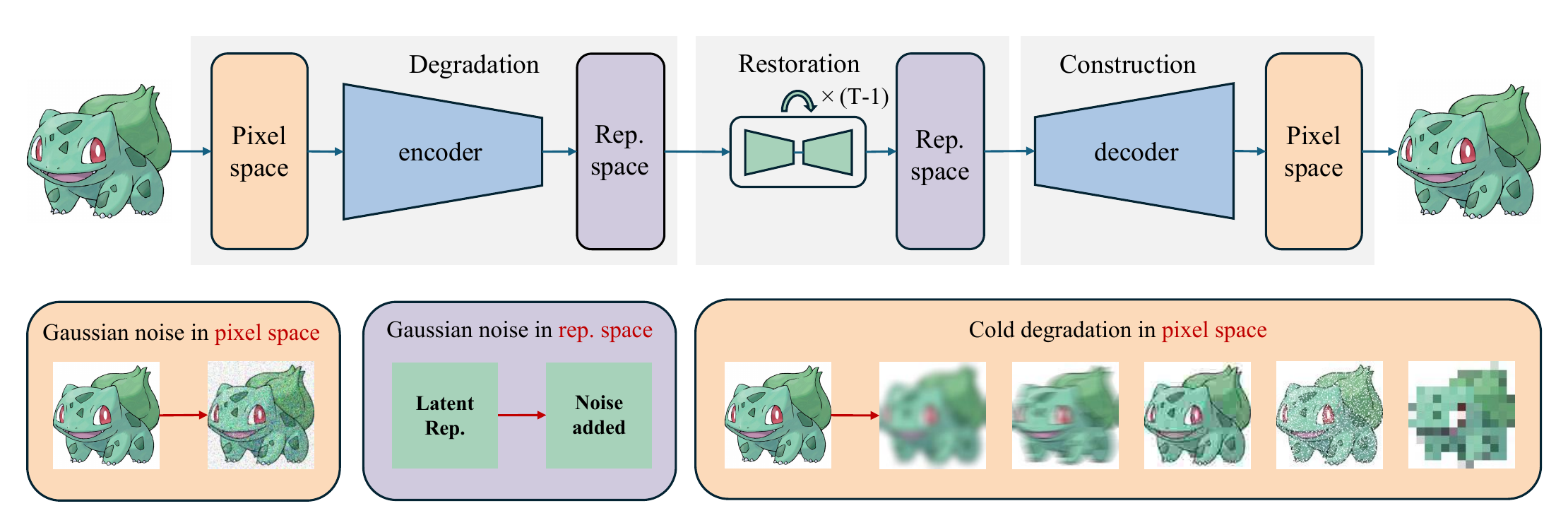} 
  \caption{The architecture of \textbf{\textit{DeAttack}}. A unified framework for watermark removal, utilizing image degradation and restoration processes.}
  \label{fig:deattack}
\end{figure*}

\begin{table*}[!t]
\centering
\setlength\tabcolsep{0.3pt}
\caption{Results of different DeAttack methods on CelebA-HQ (LoRA, w/o te). The last nine gray-shaded columns correspond to our methods. The best and worst performance are marked by \textcolor{red}{red} and \textcolor{blue}{blue}, respectively.}
\begin{adjustbox}{width=\textwidth}
\begin{tabular}{c|ccc|ccc|ccc|
>{\columncolor[gray]{0.93}}ccc|
>{\columncolor[gray]{0.93}}ccc|
>{\columncolor[gray]{0.93}}ccc}
\toprule
\multirow{2.5}{*}{\textbf{Method}} 
& \multicolumn{3}{c|}{\textbf{Bmshj2018~\cite{balle2018variational}}}
& \multicolumn{3}{c|}{\textbf{Cheng~\textit{et al}.~\cite{cheng2020learned}}}
& \multicolumn{3}{c|}{\textbf{Diffusion~\cite{zhao2024invisible}}}
& \multicolumn{3}{>{\columncolor[gray]{0.93}}c|}{\textbf{SwinIR (Denoise)}} 
& \multicolumn{3}{>{\columncolor[gray]{0.93}}c|}{\textbf{SwinIR (JPEG AR)}} 
& \multicolumn{3}{>{\columncolor[gray]{0.93}}c}{\textbf{IRNeXt (Deblur)}} \\
\cmidrule(lr){2-4} \cmidrule(lr){5-7} \cmidrule(lr){8-10} 
\cmidrule(lr){11-13} \cmidrule(lr){14-16} \cmidrule(lr){17-19}
& CLIP-T$\uparrow$ & FID$\downarrow$ & Acc.(\%)$\uparrow$
& CLIP-T$\uparrow$ & FID$\downarrow$ & Acc.(\%)$\uparrow$
& CLIP-T$\uparrow$ & FID$\downarrow$ & Acc.(\%)$\uparrow$
& \cellcolor[gray]{0.93}CLIP-T$\uparrow$ & \cellcolor[gray]{0.93}FID$\downarrow$ & \cellcolor[gray]{0.93}Acc.(\%)$\uparrow$
& \cellcolor[gray]{0.93}CLIP-T$\uparrow$ & \cellcolor[gray]{0.93}FID$\downarrow$ & \cellcolor[gray]{0.93}Acc.(\%)$\uparrow$
& \cellcolor[gray]{0.93}CLIP-T$\uparrow$ & \cellcolor[gray]{0.93}FID$\downarrow$ & \cellcolor[gray]{0.93}Acc.(\%)$\uparrow$ \\
\midrule
DIAGNOSIS 
& 0.2594 & \textcolor{blue}{273.36} & 72.00 
& 0.2613 & 283.34 & 64.00 
& 0.2628 & \textcolor{blue}{280.06} & 84.00 
& \cellcolor[gray]{0.93}0.2557 & \cellcolor[gray]{0.93}269.89 & \cellcolor[gray]{0.93}70.00 
& \cellcolor[gray]{0.93}0.2531 & \cellcolor[gray]{0.93}275.03 & \cellcolor[gray]{0.93}68.00 
& \cellcolor[gray]{0.93}0.2508 & \cellcolor[gray]{0.93}278.79 & \cellcolor[gray]{0.93}78.00 \\

DiffusionShield 
& \textcolor{red}{0.2611} & \textcolor{red}{268.64} & \textcolor{red}{99.99} 
& 0.2609 & 282.65 & \textcolor{red}{99.78} 
& 0.2654 & 274.22 & \textcolor{red}{100.00} 
& \cellcolor[gray]{0.93}\textcolor{red}{0.2614} & \cellcolor[gray]{0.93}\textcolor{red}{268.66} & \cellcolor[gray]{0.93}\textcolor{red}{98.39} 
& \cellcolor[gray]{0.93}0.2565 & \cellcolor[gray]{0.93}272.21 & \cellcolor[gray]{0.93}\textcolor{red}{100.00} 
& \cellcolor[gray]{0.93}0.2519 & \cellcolor[gray]{0.93}\textcolor{red}{257.92} & \cellcolor[gray]{0.93}\textcolor{red}{100.00} \\

WatermarkDM 
& \textcolor{blue}{0.2571} & 270.79 & 62.50 
& \textcolor{red}{0.2652} & \textcolor{red}{277.76} & 57.81
& \textcolor{red}{0.2696} & \textcolor{red}{269.35} & 54.69 
& \cellcolor[gray]{0.93}0.2589 & \cellcolor[gray]{0.93}269.68 & \cellcolor[gray]{0.93}\textcolor{blue}{51.56} 
& \cellcolor[gray]{0.93}\textcolor{red}{0.2579} & \cellcolor[gray]{0.93}\textcolor{red}{268.63} & \cellcolor[gray]{0.93}\textcolor{blue}{48.44} 
& \cellcolor[gray]{0.93}\textcolor{red}{0.2525} & \cellcolor[gray]{0.93}258.28 & \cellcolor[gray]{0.93}56.25 \\

SIREN 
& 0.2601 & 270.64 & \textcolor{blue}{51.83} 
& \textcolor{blue}{0.2593} & \textcolor{blue}{285.56} & \textcolor{blue}{52.29} 
& \textcolor{blue}{0.2554} & 277.21 & \textcolor{blue}{51.92} 
& \cellcolor[gray]{0.93}0.2584 & \cellcolor[gray]{0.93}\textcolor{blue}{273.65} & \cellcolor[gray]{0.93}52.38 
& \cellcolor[gray]{0.93}0.2555 & \cellcolor[gray]{0.93}\textcolor{blue}{275.34} & \cellcolor[gray]{0.93}52.38 
& \cellcolor[gray]{0.93}0.2519 & \cellcolor[gray]{0.93}263.98 & \cellcolor[gray]{0.93}\textcolor{blue}{52.21} \\
\bottomrule
\end{tabular}
\end{adjustbox}
\label{tab_attack}
\end{table*}

\section{Watermark Removal Approach}
\label{sec:removal}
Modern watermarking algorithms exhibit robustness against common degradations, necessitating more targeted and principled removal strategies. Existing regeneration-based methods~\cite{sun2023denet,bui2025trustmark,bansal2023cold}—including diffusion reconstruction, adversarial perturbations, and latent editing—typically disrupt the watermark by modifying the input distribution or internal representations. However, these operations are often unstructured and suboptimal.

We denote the original clean image as \(I(x,y)\), the embedded watermark as \(W(x,y)\), and the watermarked image as:
\begin{equation}
I_w(x,y) = I(x,y) + \alpha W(x,y),
\end{equation}
where \(\alpha \ll 1\) is the embedding strength. The goal of watermark removal is to estimate a clean reconstruction \(\hat{I}\) that minimizes the residual watermark energy:
\begin{equation}
\hat{I} = \arg\min_{I'} \; \| I' - I \|_2^2 + \lambda \| \mathcal{E}(I') \|_2^2,
\end{equation}
where \(\mathcal{E}(\cdot)\) denotes a watermark extractor or spectral energy operator.

\textbf{Additive Gaussian noise in pixel space.}
An attacker may inject pixel-level Gaussian noise:
\begin{equation}
I'(x,y) = I_w(x,y) + n(x,y), \quad n(x,y) \sim \mathcal{N}(0, \sigma^2).
\end{equation}
In the Fourier domain:
\begin{equation}
\hat{I}'(u,v) = \hat{I}_w(u,v) + \hat{n}(u,v),
\quad
\mathbb{E}[|\hat{n}(u,v)|^2] = \sigma^2 MN.
\end{equation}
Because \(\hat{n}(u,v)\) distributes energy uniformly across frequencies, it masks \(\hat{W}(u,v)\) in mid- and high-frequency bands. However, due to its randomness, residual structured components of \(W\) may survive under robust decoders.

\textbf{Gaussian blur in pixel space.}
Gaussian blurring applies a deterministic degradation:
\begin{equation}
I'(x,y) = (I_w * G_\sigma)(x,y),
\end{equation}
where the Gaussian kernel is given by:
\begin{equation}
G_\sigma(x,y) = \frac{1}{2\pi\sigma^2}\exp\!\left(-\frac{x^2+y^2}{2\sigma^2}\right).
\end{equation}
In the frequency domain:
\begin{equation}
\hat{I}'(u,v) = \hat{I}_w(u,v) \cdot H(u,v),
\quad
H(u,v) = e^{-2\pi^2\sigma^2(u^2+v^2)}.
\end{equation}
Here, \(H(u,v)\) acts as a low-pass filter, systematically attenuating high-frequency components of \(\hat{W}(u,v)\). The effective suppression ratio is:
\begin{equation}
r(u,v) = \frac{|\alpha \hat{W}(u,v)H(u,v)|^2}{|\alpha \hat{W}(u,v)|^2} = |H(u,v)|^2,
\end{equation}
showing that watermark energy decays exponentially as frequency increases.

\textbf{Additive Gaussian noise in latent space.}
For encoder-based models \(z = E(I)\), we apply latent perturbation:
\begin{equation}
z' = z + \epsilon, \quad \epsilon \sim \mathcal{N}(0, \sigma^2 I_d),
\end{equation}
where \(I_d\) is the \(d\times d\) identity matrix. The reconstructed image is:
\begin{equation}
I' = D(z') = D(E(I) + \epsilon).
\end{equation}
The expected reconstruction error can be approximated by:
\begin{equation}
\mathbb{E}[\|I' - I\|_2^2] \approx \sigma^2 \operatorname{Tr}(J_D J_D^\top),
\end{equation}
where \(J_D = \partial D / \partial z\) is the decoder. Since \(J_D\) emphasizes low- and mid-frequency bases, this perturbation predominantly removes smooth, semantic components but is less effective on fine-grained, high-frequency watermarks.

We generalize these operations into a unified regeneration attack:
\begin{equation}
x' = \mathcal{R}(\mathcal{D}(x)),
\end{equation}
where \(\mathcal{D}\) is degradation operator and \(\mathcal{R}\) is restoration operator, respectively.

The degradation model can be expressed as:
\begin{equation}
\mathcal{D}(x) = x + \delta, \quad \delta \sim \mathcal{P}_\theta,
\end{equation}
where \(\mathcal{P}_\theta\) defines a learnable degradation prior. Restoration reconstructs a clean estimate:
\begin{equation}
\mathcal{R}(y) = \arg\min_x \|y - x\|_2^2 + \beta \Phi(x),
\end{equation}
where \(\Phi(x)\) encodes image regularity.

We propose \textbf{\textit{DeAttack}}, shown in Figure.~\ref{fig:deattack}, a unified framework combining multi-domain degradations and high-quality restorations. Its autoencoder backbone integrates both pixel-space and latent-space degradations, followed by adaptive restoration blocks to recover perceptual quality.

Specifically, we first employ IRNeXt~\cite{cui2023irnext,cui2024revitalizing} as the core restoration network, trained on DIV2K~\cite{agustsson2017ntire}, Flickr2K, and WED~\cite{ma2016waterloo}. Gaussian-blurred samples (\(71\times71\) kernel, \(\sigma=15\)) simulate high-frequency attenuation. Next, we use two pretrained SwinIR~\cite{liang2021swinir} models are used for denoising and JPEG artifact correction, our \textbf{\textit{IRNeXt-based DeAttack}} achieves stronger watermark removal with minimal perceptual distortion, as validated in Table~\ref{tab_attack}.

\section{Conclusion}
\label{sec:conclusion}
This paper analyzes image watermarking techniques for tracing unauthorized fine-tuning in Stable Diffusion. Experimental results show these methods lack robustness in real-world scenarios. The results of some existing methods exhibit universality across diverse fine-tuning approaches and tasks, as well as transmissibility even when only a small proportion of watermarked images is used. Finally, we propose DeAttack, a unified watermark removal framework based on image degradation and restoration. We assess how various types of degradation impact watermark removal. Results show our method can inspire more powerful watermarking techniques.
\clearpage
\setcounter{page}{1}
\maketitlesupplementary
The supplementary material provides additional details that complement the main paper. It includes dataset descriptions, implementation details, extended experimental results, qualitative visualizations, and full prompt lists that could not be included in the main manuscript due to space constraints. Unless otherwise specified, all supplementary experiments follow the same evaluation protocol, metric settings, and training pipeline as those used in the main paper, in order to ensure consistency and comparability across all reported results.
\section{Dataset Details}
In this section, we provide additional information about the three datasets used in Sec. 4 of the main paper. These datasets are selected to cover three representative protection scenarios, namely facial identity protection, virtual object protection, and artistic style protection.
\begin{itemize}
    \item \textit{\textbf{CelebA}}~\cite{liu2015deep}: This dataset contains face images of celebrities. For our experiments, each image is paired with a descriptive caption generated by LLaVA~\cite{liu2023visual}. Since the original dataset is substantially larger than required for diffusion model fine-tuning in our setting, we randomly sample 1,000 image-caption pairs for experiments. This dataset is mainly used to evaluate the preservation and traceability of identity-related content.
    \item \textit{\textbf{Pokémon}}~\cite{pinkney2022pokemon}: This dataset consists of 833 high-quality Pokemon images, each accompanied by a text caption produced by the BLIP~\cite{li2022blip} captioning model. Compared with real-image benchmarks, this dataset emphasizes stylized and character-centric visual concepts, making it suitable for evaluating watermark behavior in virtual object customization settings.
    \item \textit{\textbf{WikiArt}}~\cite{wikiart}: This dataset contains 81,444 artwork images collected from WikiArt.org, with annotations such as artist, genre, and style. Owing to its broad stylistic diversity and rich artistic textures, WikiArt serves as a challenging benchmark for studying artistic style protection and the transferability of watermark-related effects under fine-tuning.
\end{itemize}
Together, these three datasets provide complementary evaluation scenarios and enable a more comprehensive analysis of watermark robustness, traceability, and removal effectiveness under different data characteristics.
\section{Experimental Settings}
All experiments are implemented in Python 3.10 with PyTorch 2.7.1, and are conducted on Ubuntu 20.04 using four NVIDIA A800 GPUs. Unless explicitly stated otherwise, the same computational environment and evaluation pipeline are used throughout all experiments.
\subsection{\textbf{Models and Datasets}}
We consider four representative dataset watermarking methods, namely DIAGNOSIS~\cite{wang2024diagnosis}, DiffusionShield~\cite{cui2025diffusionshield}, SIREN \cite{li2025towards}, and WatermarkDM~\cite{zhao2023recipe}, as the primary protection baselines. These methods embed imperceptible watermarks into the original training images, with the goal of enabling the tracing of unauthorized data usage during downstream diffusion model customization. Following the setup in the main paper, we evaluate them on three high-resolution datasets, CelebA-HQ~\cite{liu2015deep}, Pokémon~\cite{pinkney2022pokemon}, and WikiArt~\cite{wikiart}, corresponding to face protection, virtual object protection, and artistic style protection, respectively.
\subsection{\textbf{Implementation Details}}
During watermark embedding, we follow the default settings of each watermarking method and resize all images to 512 × 512. During diffusion model fine-tuning, we uniformly adopt Stable Diffusion v1.4 (SD1.4)~\cite{rombach2022high} as the default backbone and apply the same prompt template across all four fine-tuning strategies, so that the comparison is not affected by differences in prompting format. Each fine-tuning run is based on 10 training images sampled from the corresponding dataset. During evaluation, we use 50 prompts for each dataset, and generate five images per prompt. The final FID and CLIP similarity scores are reported as the average over the generated samples. This unified setup ensures that the observed performance differences mainly reflect the effects of watermarking and data removal, rather than confounding factors introduced by inconsistent fine-tuning or generation conditions.

\section{Common Distortion Processing}
To evaluate watermark robustness under realistic perturbations, we consider three commonly used natural distortions: Gaussian blur, JPEG compression, and Gaussian noise. These distortions are designed to simulate typical image degradation encountered in practical scenarios such as image storage, transmission, and post-processing.
\begin{algorithm}[!t]
\caption{Apply Image Distortions: Gaussian Blur, JPEG Compression, and Gaussian Noise}
\label{alg:distortions}
\begin{algorithmic}[1]
\Require Original image $I$; JPEG quality $q = 15$; Gaussian noise std $\sigma = 0.3$ 
\Ensure Distorted images: $I_{\text{blur}}$, $I_{\text{jpeg}}$, $I_{\text{noise}}$

\State \textbf{Gaussian Blur:}
\State $I_{\text{blur}} \gets \text{GaussianBlur}(I, \text{kernel\_size}=31, \sigma=0)$

\State \textbf{JPEG Compression:}
\State $I_{\text{jpeg}} \gets \text{JPEGEncode}(I, \text{quality}=q)$

\State \textbf{Gaussian Noise:}
\State $I_{\text{norm}} \gets I / 255$
\State Sample $\varepsilon \sim \mathcal{N}(0, \sigma^2)$
\State $I_{\text{noise}} \gets \text{clip}(I_{\text{norm}} + \varepsilon, 0, 1)$
\State $I_{\text{noise}} \gets I_{\text{noise}} \times 255$

\State \textbf{Return:} $I_{\text{blur}}, I_{\text{jpeg}}, I_{\text{noise}}$
\end{algorithmic}
\end{algorithm}
\subsection{\textbf{Image Blur}}
For Gaussian blur, we convolve the image with a two-dimensional Gaussian kernel. In our implementation, a \(31\times31\) kernel is used, and the standard deviation is automatically determined by the underlying image processing function. This distortion mainly smooths local image details and attenuates high-frequency information.
\begin{equation}
I_{\text{blur}}(x, y) = (I * G)(x, y) = \sum_{u=-k}^{k} \sum_{v=-k}^{k} I(x-u, y-v) \cdot G(u,v),
\end{equation}
\begin{equation}
G(u,v) = \frac{1}{2\pi\sigma^2} \exp\left( -\frac{u^2 + v^2}{2\sigma^2} \right),
\end{equation}
where $I$ is the original image, $G$ is a two-dimensional Gaussian kernel, $\sigma$ is standard deviation. We used a $31\times31$ kernel, which automatically calculates the corresponding standard deviation.
\subsection{\textbf{JPEG Compression}}
For JPEG compression, the image is first partitioned into \(8\times8\) blocks, transformed into the frequency domain by the discrete cosine transform (DCT), and then quantize the frequency domain coefficients $C(u,v)$, as follows:
\begin{equation}
C_q(u,v) = \text{round} \left( \frac{C(u,v)}{Q(u,v)} \right),
\end{equation}
where $Q(u,v)$ is the standard quantization matrix. The distortion mainly comes from this quantization operation. The JPEG quality level used in the code is 15.
\subsection{\textbf{Gaussian Noise}}
For Gaussian noise, zero-mean Gaussian noise is independently added to each color channel, followed by clipping the perturbed pixel values to the valid range, the process is as follows:
\begin{equation}
I_{\text{noisy}}(x, y, c) = \text{clip} \left( I(x, y, c) + \mathcal{N}(0, \sigma^2), 0, 1 \right),
\end{equation}
where $\mathcal{N}(0, \sigma^2)$ represents Gaussian noise with mean $0$ and variance $x$; the $clip$ function clamps the pixel values to the interval $[0,1]$. The distortion is added independently in the dimension of the color channel $c$.

The complete distortion pipeline is summarized in Algorithm~\ref{alg:distortions}, which provides a unified description of the three perturbation processes used in our experiments.

\begin{table*}[!t]
    \centering
    \caption{Experimental results of SDXL.}
    \begin{tabular}{c|ccc|ccc|ccc}
        \toprule
        \multirow{1}{*}{\makecell{FT\\Method}} & \multicolumn{3}{c}{CelebA-HQ} & \multicolumn{3}{c}{Pokémon} & \multicolumn{3}{c}{WikiArt} \\
        \cmidrule(lr){2-10}
        & CLIP-T$\uparrow$ & FID$\downarrow$ & Acc.$\uparrow$ & CLIP-T$\uparrow$ & FID$\downarrow$ & Acc.$\uparrow$ & CLIP-T$\uparrow$ & FID$\downarrow$ & Acc.$\uparrow$ \\
        \midrule
        T2I & 0.1771 & 284.35 & \textcolor{red}{96.35} & 0.1915 & 264.73 & \textcolor{red}{98.44} & 0.1137 & 374.39 & 70.32 \\
        LoRA & \textcolor{red}{0.2287} & \textcolor{red}{263.33} & 90.32 & 0.2106 & \textcolor{red}{216.57} & 92.83 & 0.1324 & 378.41 & 70.21 \\
        DB & 0.2249 & 285.64 & 88.73 & \textcolor{red}{0.2559} & 279.40 & 92.28 & \textcolor{red}{0.2232} & \textcolor{red}{350.25} & \textcolor{red}{83.36} \\
        TI & 0.1917 & 291.32 & 92.59 & 0.2437 & 243.59 & 90.35 & 0.2139 & 360.78 & 81.59 \\
        \bottomrule
    \end{tabular}
    \label{tab:other_diffusion}
\end{table*}
\begin{table*}[!t]
    \centering
    \caption{Results of DeAttack on SDXL.}
    \begin{tabular}{c|ccc|ccc|ccc}
        \toprule
        \multirow{1}{*}{\makecell{FT\\Method}} & \multicolumn{3}{c}{CelebA-HQ} & \multicolumn{3}{c}{Pokémon} & \multicolumn{3}{c}{WikiArt} \\
        \cmidrule(lr){2-10}
        & CLIP-T$\uparrow$ & FID$\downarrow$ & Acc.$\uparrow$ & CLIP-T$\uparrow$ & FID$\downarrow$ & Acc.$\uparrow$ & CLIP-T$\uparrow$ & FID$\downarrow$ & Acc.$\uparrow$ \\
        \midrule
        T2I & 0.1934 & 284.23 & 52.59 & 0.1972 & 272.49 & \textcolor{red}{54.12} & 0.1359 & \textcolor{red}{343.74} & \textcolor{red}{58.89} \\
        LoRA & \textcolor{red}{0.2389} & \textcolor{red}{254.89} & 51.52 & 0.2109 & 296.37 & 49.45 & 0.1994 & 373.15 & 55.57 \\
        DB & 0.2322 & 258.71 & \textcolor{red}{58.73} & \textcolor{red}{0.2775} & \textcolor{red}{234.31} & 54.10 & 0.2021 & 354.69 & 56.74 \\
        TI & 0.2301 & 302.29 & 55.37 & 0.2697 & 265.91 & 49.86 & \textcolor{red}{0.2103} & 356.98 & 57.21 \\
        \bottomrule
    \end{tabular}
    \label{tab:other_diffusion_deattack}
\end{table*}

\section{Experimental Analysis}
In the main paper, we adopt Stable Diffusion v1.4 (SD1.4) as the default diffusion backbone, following the evaluation setting commonly used in prior watermarking studies. This choice is also consistent with the configuration adopted by most mainstream baselines, which facilitates fair comparison under a unified experimental protocol. To examine whether the conclusions of this work extend beyond a single diffusion architecture, we further conduct supplementary experiments on SDXL. The results reported in Table~\ref{tab:other_diffusion} show that the overall benchmark observations remain consistent when moving from SD1.4 to a stronger backbone, indicating that the conclusions drawn in the main paper are not specific to SD1.4. In particular, the relative performance trends across different fine-tuning strategies are largely preserved, suggesting that the benchmark captures stable properties of dataset watermarking rather than artifacts of a particular model choice.

We further report the performance of DeAttack on SDXL in Table~\ref{tab:other_diffusion_deattack}. The results show that DeAttack remains effective under a different diffusion architecture, demonstrating that its removal capability is not tightly coupled to the original SD1.4 setting. This observation suggests that the method has reasonable architectural generalizability and can transfer to stronger generation backbones without substantial degradation in effectiveness.

To further assess the robustness of DeAttack, we additionally evaluate it on multiple datasets and under different fine-tuning strategies, with the corresponding results presented in Table~\ref{tab:dataset_deattack}. The overall trends remain consistent with those observed in the main paper, indicating that the effectiveness of DeAttack is not limited to a single dataset or a specific fine-tuning configuration. Instead, the method exhibits stable behavior across different data distributions and customization settings, which further supports its practical applicability.

In addition, the main paper reports experimental statistics under different data protection ratios, where the clean setting corresponds to a protection ratio of 0\%. Since this condition already appears in the main paper as the unprotected reference case, we further provide in Table~\ref{tab:distortion_clean} the results of natural distortion under the clean setting for completeness. These results offer a clearer reference point for understanding model behavior in the absence of watermark protection, and they facilitate comparison with the corresponding results under non-zero protection ratios.
\begin{table*}[!t]
    \centering 
    \caption{Results of DeAttack on three datasets and fine-tuning method.}
    \begin{tabular}{c|ccc|ccc|ccc}
        \toprule
        \multirow{3}{*}{\makecell{FT\\Method}} & \multicolumn{3}{c}{CelebA-HQ} & \multicolumn{3}{c}{Pokémon} & \multicolumn{3}{c}{WikiArt} \\
        \cmidrule(lr){2-10}
        & CLIP-T$\uparrow$ & FID$\downarrow$ & Acc.$\uparrow$ & CLIP-T$\uparrow$ & FID$\downarrow$ & Acc.$\uparrow$ & CLIP-T$\uparrow$ & FID$\downarrow$ & Acc.$\uparrow$ \\
        \midrule
        T2I & 0.2018 & 286.39 & 54.84 & 0.1934 & 265.37 & 53.10 & 0.1090 & 362.25 & \textcolor{red}{56.76} \\
        DB & \textcolor{red}{0.2589} & \textcolor{red}{253.51} & \textcolor{red}{59.86} & \textcolor{red}{0.2931} & \textcolor{red}{233.59} & \textcolor{red}{55.56} & 0.2280 & 356.20 & 54.74 \\
        TI & 0.2503 & 292.30 & 56.25 & 0.2901 & 268.81 & 50.86 & \textcolor{red}{0.2581} & \textcolor{red}{348.24} & 52.78 \\
        \bottomrule
    \end{tabular}
    \label{tab:dataset_deattack}
\end{table*}
\begin{table*}[!t]
    \centering
    \caption{Results of clean data under different natural distortion.}
    \begin{adjustbox}{width=\textwidth}
    \begin{tabular}{c|ccc|ccc|ccc|ccc}
        \toprule
        \multirow{2.5}{*}{\makecell{Distortion\\Type}} & \multicolumn{3}{c}{Text-to-Image} & \multicolumn{3}{c}{LoRA} & \multicolumn{3}{c}{DreamBooth} & \multicolumn{3}{c}{Textual Inversion} \\
        \cmidrule(lr){2-13}
        & CLIP-T$\uparrow$ & FID$\downarrow$ & Acc.$\uparrow$ & CLIP-T$\uparrow$ & FID$\downarrow$ & Acc.$\uparrow$ & CLIP-T$\uparrow$ & FID$\downarrow$ & Acc.$\uparrow$ & CLIP-T$\uparrow$ & FID$\downarrow$ & Acc.$\uparrow$ \\
        \midrule
        Blur(w/o) & \textcolor{red}{0.2274} & 230.81 & N/A & \textcolor{red}{0.2621} & \textcolor{red}{241.74} & N/A & \textcolor{red}{0.2558} & \textcolor{red}{223.45} & N/A & \textcolor{red}{0.2617} & 243.79 & N/A \\
        Blur(w) & 0.2150 & \textcolor{red}{217.95} & N/A & 0.2412 & 273.47 & N/A & 0.2533 & 240.02 & N/A & 0.2489 & \textcolor{red}{239.61} & N/A \\
        JPEG(w/o) & 0.2175 & 235.92 & N/A & 0.2580 & 243.39 & N/A & 0.2512 & 232.12 & N/A & 0.2594 & 242.72 & N/A \\
        JPEG(w) & 0.2163 & 220.83 & N/A & 0.2386 & 278.88 & N/A & 0.2513 & 244.59 & N/A & 0.2502 & 241.45 & N/A \\
        Noise(w/o) & 0.1975 & 240.36 & N/A & 0.2039 & 275.44 & N/A & 0.2201 & 236.86 & N/A & 0.2332 & 252.81 & N/A \\
        Noise(w) & 0.1992 & 229.62 & N/A & 0.2014 & 279.23 & N/A & 0.2201 & 269.35 & N/A & 0.2293 & 243.89 & N/A \\
        \bottomrule
    \end{tabular}
    \end{adjustbox}
    \label{tab:distortion_clean}
\end{table*}
\section{Visualization}
\subsection{Frequency domain distribution}
\begin{figure*}[!th]
  \centering
  \includegraphics[width=.9\textwidth]{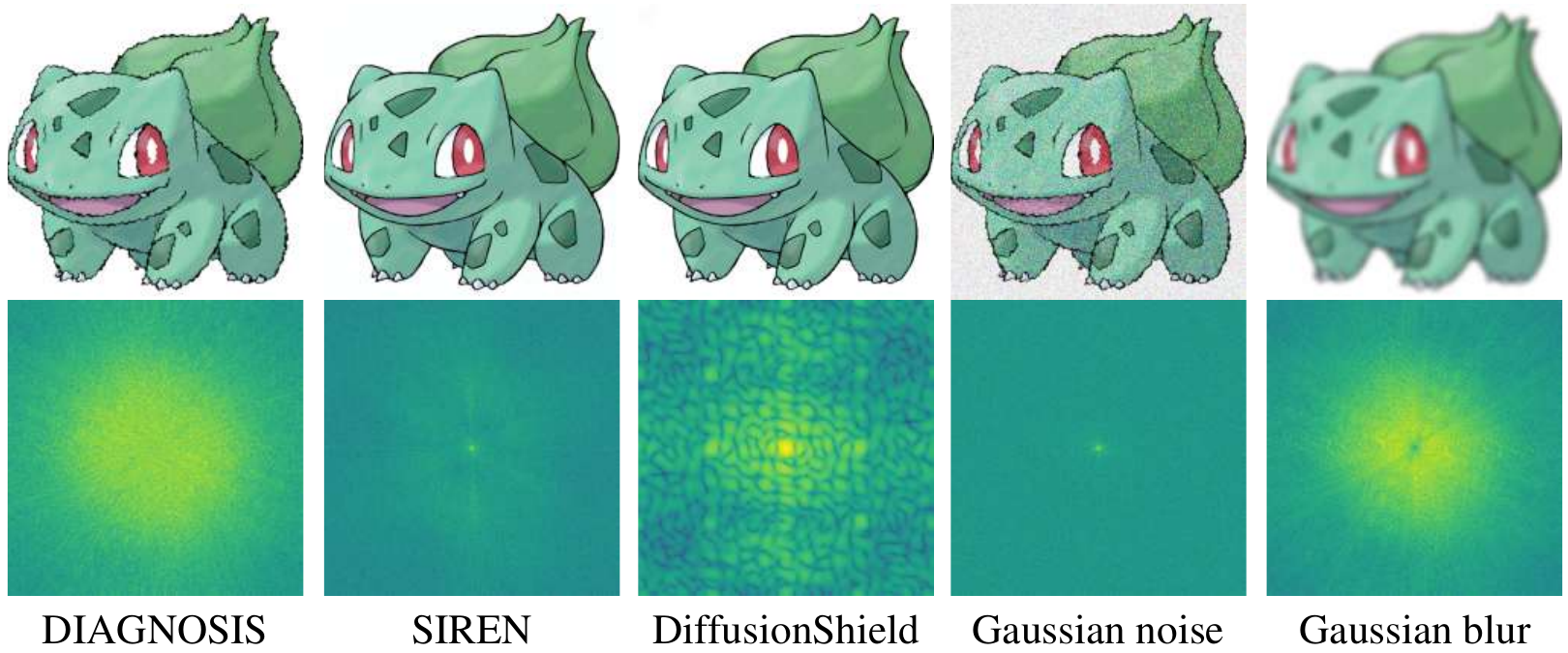} 
  \caption{Heatmap visualization of frequency-domain changes caused by watermarking methods and natural distortion.}
  \label{fig:deattack_vis}
\end{figure*}
This section analyzes how different watermarking methods, as well as natural distortions, affect the frequency-domain characteristics of the original image. As shown in Figure~\ref{fig:deattack_vis}, we visualize the perturbation patterns introduced by DIAGNOSIS~\cite{wang2024diagnosis}, SIREN~\cite{li2025towards}, and DiffusionShield~\cite{cui2025diffusionshield}, Gaussian noise, and Gaussian blur using heatmaps in the Fourier domain. Specifically, the Fourier transform is applied to project pixel-level perturbations into the frequency space, where brighter yellow regions indicate larger-magnitude deviations and greener regions indicate relatively smaller changes.

From a spatial perspective, the center of the spectrum corresponds to low-frequency components, while regions farther from the center correspond to higher-frequency components. Based on this visualization, different watermarking methods exhibit distinct spectral characteristics. DIAGNOSIS tends to produce stronger perturbations concentrated in the low-frequency region, whereas DiffusionShield shows a comparatively more distributed response across both low- and high-frequency bands. This suggests that different watermarking methods encode traceability signals in different spectral patterns, which may partly explain differences in robustness and removability observed in the main experiments.

The two natural distortions also exhibit recognizable frequency-domain behaviors. Gaussian noise introduces dispersed perturbations, while Gaussian blur leads to more pronounced low-frequency changes and stronger degradation in the overall spectrum. Compared with Gaussian noise, Gaussian blur produces more visually concentrated and larger-magnitude changes, indicating that it may interfere more strongly with the spectral structure of the original image. Overall, these visualizations provide an intuitive explanation of how watermark-induced perturbations differ from natural distortions, and help clarify why some watermarking methods are more resilient or more vulnerable under certain post-processing operations.
\begin{figure*}[!t]
  \centering
  \includegraphics[width=.72\textwidth]{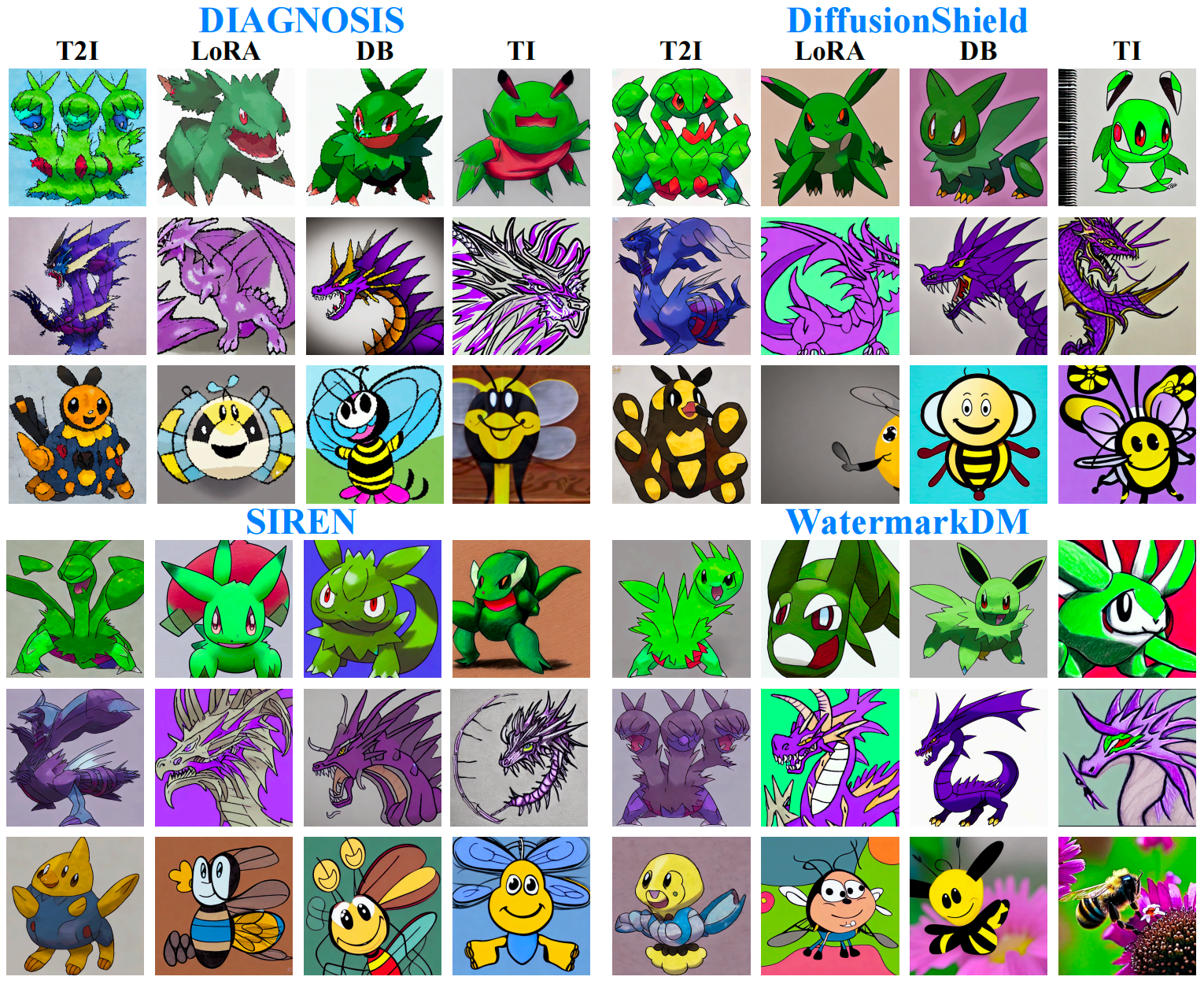} 
  \caption{Pokémon dataset fine-tuning results.}
  \label{fig:finetunepokemon}
\end{figure*}
\begin{figure*}[!t]
  \centering
  \includegraphics[width=.72\textwidth]{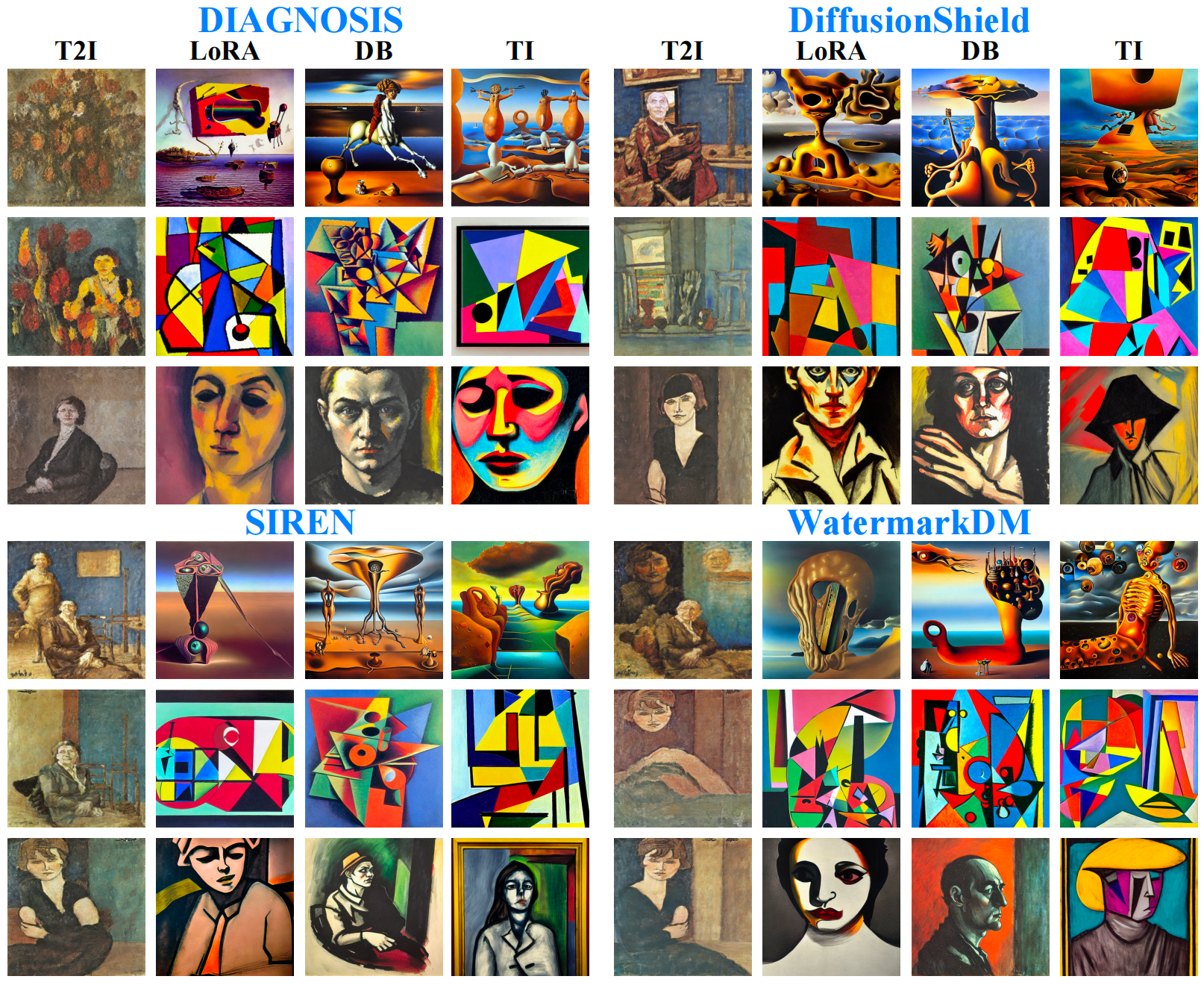} 
  \caption{WikiArt dataset fine-tuning results.}
  \label{fig:finetunewikiart}
\end{figure*}
\subsection{Fine-tuning generation}
This section presents qualitative examples of images generated after fine-tuning on watermarked data, with comparisons across four watermarking methods and four fine-tuning strategies. The purpose of these visualizations is to provide an intuitive view of how watermark embedding influences downstream generation behavior under different customization settings.

As shown in Figure~\ref{fig:finetunepokemon}, the examples on the Pokemon dataset highlight the impact of watermarking in a stylized, character-centric domain. Since Pokemon images typically contain clear contours, saturated colors, and relatively simple semantic compositions, visual differences across methods can be more directly observed in object appearance, color consistency, and structural fidelity. In contrast, Figure~\ref{fig:finetunewikiart} presents results on the WikiArt dataset, which contains diverse artistic styles, painterly textures, and more complex visual abstractions. In this setting, the influence of watermarking is reflected not only in object-level structure, but also in style transfer behavior, brushstroke patterns, and texture coherence.

Taken together, these qualitative comparisons show that the visual effects of watermark embedding and removal are not limited to a single visual domain. Instead, they can be observed across both structured cartoon-like images and highly diverse artistic images. This further supports the generality of the benchmark and provides complementary evidence to the quantitative results reported in the main paper and the supplementary tables.
\section{Prompts for generation}
This section provides the complete set of textual prompts used for image generation after fine-tuning the diffusion model. For each dataset, we use 50 prompts designed to cover a broad range of semantic content and stylistic variations, so that the evaluation is not biased toward a narrow subset of concepts. More specifically, the prompts for CelebA emphasize facial attributes, accessories, and appearance-related descriptions; the prompts for Pokemon focus on character identity, shape, color, and cartoon-style object descriptions; and the prompts for WikiArt cover diverse artistic genres, painting styles, historical aesthetics, and compositional patterns.
\begin{table*}[t]
\centering
\caption{List of CelebA dataset prompts.}
\label{tab_celeba50}
\small
\setlength{\tabcolsep}{4pt}
\begin{tabular}{c|p{15.5cm}}
\toprule
\textbf{ID} & \textbf{Prompt Description} \\
\midrule
1 & A young woman with red hair, heart-shaped face, small nose, large brown eyes, glasses, and a necklace. Likely a young adult. \\
2 & A young bald man with a beard and round face, wearing a football helmet. He has thick lips and a large nose. \\
3 & A blonde young woman with a heart-shaped face, small nose, thin lips, wearing a black dress and a necklace. Smiling. \\
4 & A young woman with dark hair, heart-shaped face, full lips, straight nose, and a necklace. Likely in her late teens. \\
5 & A bald man with glasses, wide face, thick lips, wearing a suit and standing at a microphone. He is an adult. \\
6 & An elderly white man with glasses, wide face, large nose, thin mouth, and beard. Wearing a tie and brown jacket. \\
7 & A young blonde woman with glasses, small nose, wide mouth, brown eyes, a bracelet, and a nose piercing. No facial hair. \\
8 & A smiling young man with a round face, glasses, beard, brown eyes, and wearing a suit and tie. A flag is behind him. \\
9 & A man with a wide face, thick mustache, black hair, and glasses. Smiling, with no other accessories visible. \\
10 & A young woman with dark hair, brown cat-like eyes, heart-shaped face, wide mouth, small nose, and black dress. \\
11 & A man with blonde hair, wide face, small nose, thick lips, large eyes, and glasses. Wearing a white shirt and smiling. \\
12 & A young woman with a narrow heart-shaped face, dark hair, large eyes, small nose, thin lips, wearing pink dress and necklace. \\
13 & A smiling young blonde woman with heart-shaped face, brown eyes, small nose, glasses, thick eyebrows, and a necklace. \\
14 & A young man with glasses, dark hair, large nose, black eyes, strong jawline, and straight mouth. Likely a young adult. \\
15 & A woman with blonde bobbed hair, red dress, heart-shaped face, small nose, smiling. Possibly young, not clearly elderly. \\
16 & A young blonde woman with large expressive eyes, small nose, thin mouth, glasses, white shirt, and heart-shaped face. \\
17 & A smiling young blonde woman with heart-shaped face, blue eyes, glasses, small wide nose, thick lips, and jewelry. \\
18 & A young woman with brown eyes, glasses, thick lips, pearl necklace, heart-shaped face, small nose, and a smile. \\
19 & A young woman with auburn hair, glasses, brown eyes, wide mouth, heart-shaped face, and a necklace. \\
20 & A young woman with a ponytail, dark hair, glasses, small nose, full mouth, black clothes, and heart-shaped face. \\
21 & An elderly bald man with a beard, wide thick face, large nose, wearing a jacket and hat. Seated in front of a camera. \\
22 & A young man with shaved sides and ponytail, large brown eyes, wide upturned nose, glasses, beard, and rectangular face. \\
23 & A smiling young blonde woman with heart-shaped face, pointed nose, full lips, earrings, and brown eyes. No glasses. \\
24 & A young blonde woman with wide face, small wide nose, thick lips, large earrings, necklace, and a smile. \\
25 & A young bald man with glasses, beard, large nose, wide face, thick eyebrows, and a black jacket. Likely young adult. \\
26 & A goofy young man with round face, large eyes, small nose, thin mouth, glasses, and a playful smile. \\
27 & A young man with shaved head, large nose, wide open eyes, black hoodie, wide mouth. Possibly a teen or young adult. \\
28 & A young woman in pink dress with round face, glasses, pink bow, large brown eyes, holding a cherry in her mouth. \\
29 & An elderly bald man with glasses, large nose, thick mustache, red-striped shirt, tie, and a big smile. \\
30 & A man with long hair, beard, glasses, large nose, wide mouth, wearing a black shirt and jacket. Likely adult. \\
31 & A young blonde woman with heart-shaped face, large brown eyes, small nose, necklace, and a thin mouth. Wearing a dress. \\
32 & A smiling woman with almond-shaped eyes, wide nose, large mouth, pink dress, blonde hair. Likely a young adult. \\
33 & A young woman with red shirt, heart-shaped face, large dark eyes, full lips, small nose, necklace, and ponytail. \\
34 & A smiling young woman with heart-shaped face, small nose, large brown eyes, pink bathing suit, and pink headband. \\
35 & A young man in a suit and tie with round face, small nose, brown eyes, glasses, beard, and a thin mouth. \\
36 & A young woman with long black hair, glasses, small nose, heart-shaped face, necklace, and a thin mouth. Likely teen. \\
37 & A smiling young woman with heart-shaped face, wide mouth, large eyes, necklace, and ponytail. Possibly young adult. \\
38 & An elderly person with wide face, large black eyes, round nose, bushy eyebrows, suit, tie, and hat. \\
39 & A smiling young woman with curly dark hair, large brown eyes, full lips, small nose, necklace, and heart-shaped face. \\
40 & A man with beard, sunglasses, black suit and tie, large nose, wide mouth, and prominent chin. Handsome appearance. \\
41 & An elderly bald man with white beard, glasses, very wide face, small nose, thick mouth, wearing a suit and tie. \\
42 & A woman with round face, glasses, blonde hair, thick lips, blue shirt, small wide nose, and a friendly smile. \\
43 & A man with shaved head and beard, blue shirt, wide mouth, large nose, small blue eyes, and a youthful appearance. \\
44 & A young woman with dark straight hair, glasses, brown eyes, small nose, full lips, necklace, bracelet, and oval face. \\
45 & A thin-faced bald man with glasses, large nose, close-set eyes, suit and tie, looking directly at the camera. \\
46 & A young woman with heart-shaped face, dark hair, large expressive eyes, full mouth, small nose, and earrings. \\
47 & A young woman with blue hair, red dress, large round eyes, thick lips, wide face, necklace, and blue eyes. \\
48 & A person with long black hair, white shirt, large black eyes, wide nose, glasses, and nose piercing. Teen or adult. \\
49 & A smiling young blonde woman with round face, small nose, blue eyes, glasses, necklace, and red background. \\
50 & A young man with round face, glasses, full beard, small nose, wearing a suit and tie. Appears formally dressed. \\
\bottomrule
\end{tabular}
\end{table*}
\clearpage
\begin{table*}[t]
\centering
\caption{List of Pokémon dataset prompts.}
\label{tab_Pokémon50}
\small
\setlength{\tabcolsep}{4pt}
\begin{tabular}{c|p{15.5cm}}
\toprule
\textbf{ID} & \textbf{Prompt Description} \\
\midrule
1 & a drawing of a green pokemon with red eyes \\
2 & a cartoon monkey flying with a bone in its mouth \\
3 & a drawing of a purple dragon with spikes on it's head \\
4 & a drawing of a cat sitting on top of a flower \\
5 & a pink bird with orange eyes and a pink tail \\
6 & a cartoon bee with a big smile on it's face \\
7 & a blue cartoon character with a target in his hand \\
8 & a cartoon bird with a green leaf on its head \\
9 & a drawing of a blue dinosaur with wings \\
10 & a drawing of a green pokemon sitting on top of a leaf \\
11 & a very cute looking pokemon type \\
12 & a drawing of a shark with its mouth open \\
13 & a cartoon character with a mushroom on his head \\
14 & a drawing of a cat wearing a helmet \\
15 & a drawing of a cat with a pink tail \\
16 & a cartoon elephant with a red nose and orange ears \\
17 & a drawing of a black and white animal with horns \\
18 & a drawing of a purple and white animal \\
19 & a drawing of a red and yellow insect \\
20 & a drawing of a green and yellow lizard \\
21 & a drawing of a blue and orange pokemon \\
22 & a drawing of a gray and white pokemon \\
23 & a cartoon picture of a green vegetable with eyes \\
24 & a drawing of a green cartoon character with a sad look \\
25 & a cartoon giraffe with a ball in its mouth \\
26 & a cartoon bird with a hat on its head \\
27 & a cartoon dog is standing in a pose \\
28 & a drawing of a star with a red eye \\
29 & a cartoon turtle with a tree on its back \\
30 & a drawing of a pink cartoon character \\
31 & a drawing of a fox with wings on it's back \\
32 & a blue and yellow cartoon character with its mouth open \\
33 & a cartoon mouse with a pink shirt and tie \\
34 & a cartoon character with a yellow shirt and blue pants \\
35 & a drawing of a fish with a horn on it's head \\
36 & a drawing of a white and red pokemon \\
37 & a drawing of a blue fish with yellow eyes \\
38 & a cartoon bunny flying through the air \\
39 & a drawing of a small animal with a pink nose \\
40 & a blue and white cartoon character flying through the air \\
41 & a green and yellow toy with a red nose \\
42 & a drawing of a woman in a pink dress with a dragon head \\
43 & a cartoon character with a magnifying glass \\
44 & a drawing of a blue sea turtle holding a rock \\
45 & a cartoon bear with a ring around its neck \\
46 & a cartoon cat is holding onto a leash \\
47 & a cartoon rat with its mouth open and it's mouth wide open \\
48 & a green bird with a red tail and a black nose \\
49 & a cartoon sheep is kicking a soccer ball \\
50 & a close up of a cartoon character with big eyes \\
\bottomrule
\end{tabular}
\end{table*}
\clearpage
\begin{table*}[t]
\centering
\caption{List of WikiArt dataset prompts.}
\label{tab_WikiArt50}
\small
\setlength{\tabcolsep}{4pt}
\begin{tabular}{c|p{15.5cm}}
\toprule
\textbf{ID} & \textbf{Prompt Description} \\
\midrule
1 & surreal oil painting, Salvador Dalí style, hyper-detailed, high quality \\
2 & romantic landscape, 19th-century French painting, soft brushwork, ultra high-res \\
3 & cubist still life, abstract geometric shapes, Picasso-inspired, vibrant colors \\
4 & impressionist river scene, vivid brush strokes, Claude Monet style, realistic lighting \\
5 & art nouveau floral pattern, elegant flowing lines, Alphonse Mucha inspired, intricate details \\
6 & German expressionist portrait, emotional color palette, dramatic, cinematic lighting \\
7 & Russian avant-garde constructivist poster, vintage style, bold typography, clean vector \\
8 & hyper-realistic Baroque portrait, dramatic chiaroscuro, Rembrandt style, 8K \\
9 & abstract color field painting, Rothko inspired, vivid colors, minimalist \\
10 & Italian Renaissance fresco, mythological figures, high detail, realistic faces \\
11 & minimalist geometric abstraction, Malevich style, pure shapes, modern design \\
12 & medieval illuminated manuscript, gold leaf, intricate patterns, ancient calligraphy \\
13 & gothic cathedral interior, stained glass, atmospheric light, photorealistic \\
14 & Japanese woodblock print, Hokusai style, traditional ukiyo-e, fine linework \\
15 & fauvist landscape, intense color contrasts, Matisse inspired, expressive painting \\
16 & surreal dreamscape, Magritte style, hyper-realistic, conceptual art \\
17 & art deco poster, glamorous 1920s woman, vintage illustration, high detail \\
18 & Russian symbolist painting, mystical, ethereal lighting, rich textures \\
19 & rococo palace interior, pastel colors, ornate details, photorealistic \\
20 & Dutch golden age still life, flowers and fruits, realistic lighting, master painting \\
21 & pre-Raphaelite portrait, medieval-inspired, flowing hair, detailed textile \\
22 & Chinese ink landscape, shan shui style, misty mountains, traditional painting \\
23 & Bauhaus modernist architectural drawing, clean lines, geometric composition \\
24 & Italian futurist cityscape, motion blur, dynamic angles, vibrant \\
25 & Byzantine mosaic, religious icon, gold tesserae, intricate details \\
26 & Spanish romantic painting, dramatic history scene, vivid brushwork, realistic \\
27 & symbolist fantasy scene, allegorical figures, mystical atmosphere, high detail \\
28 & social realism mural, workers, propaganda style, bold colors, large format \\
29 & abstract expressionist painting, chaotic brushstrokes, Pollock style, large canvas \\
30 & Venetian rococo carnival scene, masked figures, ornate costumes, detailed \\
31 & French rococo pastoral painting, elegant people, romantic light, high realism \\
32 & Russian lubok folk art, storytelling style, bright colors, naive art \\
33 & Egyptian revival decorative motif, hieroglyphs, ancient style, symmetrical pattern \\
34 & neoclassical sculpture study, idealized human figure, marble texture, photorealistic \\
35 & academic classical painting, mythological subject, realistic anatomy, dramatic light \\
36 & Neue Sachlichkeit portrait, German realism, neutral colors, intense gaze \\
37 & surrealist collage, Max Ernst style, dreamlike, high-res details \\
38 & pre-Columbian inspired pattern, tribal geometric symbols, earthy colors \\
39 & gothic illuminated manuscript page, ornate borders, medieval style, hyper-detailed \\
40 & classical Greek vase painting, heroic myth scene, terracotta style, authentic \\
41 & romantic seascape, stormy sky, 19th-century painting style, high detail \\
42 & Renaissance-inspired religious altarpiece, golden halos, realistic faces, dramatic \\
43 & art brut, outsider art style, raw brushstrokes, expressive emotion \\
44 & Victorian fairy painting, delicate wings, flower garden, high detail \\
45 & expressionist cityscape, angular architecture, dramatic colors, thick brush strokes \\
46 & post-impressionist village scene, vivid colors, Van Gogh style, swirling strokes \\
47 & orientalist painting, Middle Eastern architecture, rich textures, historical \\
48 & pop art reinterpretation, classical sculpture, bright bold colors, high contrast \\
49 & suprematist non-objective composition, simple shapes, modernist, clean vector \\
50 & primitivist figure painting, tribal inspiration, earthy colors, simplified forms \\
\bottomrule
\end{tabular}
\end{table*}
\clearpage
By using a relatively diverse prompt set for each domain, we aim to better evaluate how watermarking and watsermark removal affect the generation quality, semantic alignment, and style preservation of fine-tuned diffusion models. The full prompt lists are reported in Table~\ref{tab_celeba50}, Table~\ref{tab_Pokémon50}, and Table~\ref{tab_WikiArt50}, respectively, to facilitate reproducibility and future comparison.
{
    \small
    \bibliographystyle{ieeenat_fullname}
    \bibliography{main}
}

\end{document}